%% file: main.tex
\documentclass[10pt,twocolumn,letterpaper]{article}

\usepackage{iccv}
\usepackage{times}
\usepackage{epsfig}
\usepackage{graphicx}
\usepackage{amsmath}
\usepackage{amssymb}
\usepackage{booktabs}
\usepackage{xcolor}
\usepackage[caption=false]{subfig}

\usepackage{algorithm}
\usepackage[noend]{algpseudocode}
\usepackage{eqnarray}
\usepackage{array}
\usepackage{relsize}
\usepackage[page]{appendix}

\input{macros}

\usepackage[pagebackref=true,breaklinks=true,letterpaper=true,colorlinks,bookmarks=false]{hyperref}

\iccvfinalcopy 

\begin{document}

\title{Learning Target Candidate Association to Keep Track of What Not to Track}

\newcommand{\asep}{\hspace{6mm}}
\newcommand{\aand}{\hspace{5mm}}
\author{Christoph Mayer \aand Martin Danelljan \aand Danda Pani Paudel \aand Luc Van Gool\vspace{2mm}\\
Computer Vision Lab, D-ITET, ETH Z\"urich, Switzerland
}

\maketitle

\begin{abstract}

The presence of objects that are confusingly similar to the tracked target, poses a fundamental challenge in appearance-based visual tracking. 
Such \emph{distractor objects} are easily misclassified as the target itself, leading to eventual tracking failure.
While most methods strive to suppress distractors through more powerful appearance models, we take an alternative approach. 

We propose to \emph{keep track} of distractor objects in order to continue tracking the target. 
To this end, we introduce a learned association network, allowing us to propagate the identities of all target candidates from frame-to-frame.
To tackle the problem of lacking ground-truth correspondences between distractor objects in visual tracking, we propose a training strategy that combines partial annotations with self-supervision.
We conduct comprehensive experimental validation and analysis of our approach on several challenging datasets.
Our tracker sets a new state-of-the-art on six benchmarks, achieving an AUC score of $67.1\%$ on LaSOT~\cite{Fan_2019_CVPR_Lasot} and a $+5.8\%$ absolute gain on the OxUvA long-term dataset~\cite{Valmadre_2018_ECCV_OxUvA}. The code and trained models are available at \mbox{\url{https://github.com/visionml/pytracking}}
\end{abstract}

\input{sections/introduction}
\input{sections/relatedwork}
\input{sections/method}

\input{sections/experiments_and_results}
\input{sections/discussion_and_conclusion}

{
\noindent\textbf{Acknowledgments:}
This work was partly supported by the ETH Z\"urich Fund (OK), Siemens Smart Infrastructure, the ETH Future Computing Laboratory (EFCL) financed by a gift from Huawei Technologies, an Amazon AWS grant, and an Nvidia hardware grant.}

{\small
\bibliographystyle{ieee_fullname}
\bibliography{references}
}

\clearpage
\begin{appendices}
    \input{supplementary/supplementary}
\end{appendices}

\end{document}

%% file: macros.tex
\usepackage{bm}
\usepackage{colortbl}
\usepackage{scalerel,amssymb}
\usepackage{wasysym}

\newcommand{\parsection}[1]{\vspace{0.5mm}\noindent\textbf{#1:}~}

\newcommand*\makeSet[1]{\mathcal{#1}}

\renewcommand{\vec}[1]{\mathbf{#1}}

\renewcommand\matrix[1]{\bm{#1}}

\definecolor{_yellow}{RGB}{255,217,50}
\definecolor{_darkblue}{RGB}{1,25,147}
\definecolor{_red}{RGB}{238,34,12}
\definecolor{_pink}{RGB}{255,64,255}
\definecolor{_orange}{RGB}{255,147,0}
\definecolor{_blue}{RGB}{4,51,255}
\definecolor{_green}{RGB}{97,216,54}
\definecolor{_purple}{RGB}{122,129,255}
\definecolor{_skyblue}{RGB}{86,193,255}
\definecolor{_coral}{RGB}{255,100,78}
\definecolor{_violet}{RGB}{148,55,250}

\newcommand{\object}[1]{$\begingroup\color{#1}\CIRCLE\endgroup$}
\newcommand{\bbox}[1]{$\begingroup\color{#1}\blacksquare\endgroup$}

%% file: sections/introduction.tex
\section{Introduction}

Generic visual object tracking is one of the fundamental problems in computer vision. The task involves estimating the state of the target object in every frame of a video sequence, given only the initial target location. 
Most prior research has been devoted to the development
of robust appearance models, used for locating the target object in each frame.
The two currently dominating paradigms are Siamese networks \cite{Bertinetto_2016_ECCVW_SiameseFC,Li_2018_CVPR_SiamRPN,Li_2019_CVPR_SiamRPN++} and discriminative appearance modules \cite{Bhat_2019_ICCV_DIMP,Danelljan_2017_CVPR_ECO}. While the former employs a template matching in a learned feature space, the latter constructs an appearance model through a discriminative learning formulation.
Although these approaches have demonstrated promising performance in recent years, they are effectively limited by the quality and discriminative power of the appearance model. 

As one of the most challenging factors, co-occurrence of distractor objects similar in appearance to the target is a common problem in real-world tracking applications~\cite{Bhat_2020_ECCV_KYS,Zhu_2018_ECCV_DaSiamRPN,Xiao_2016_ECCV_DistractorSupportedST}. Appearance-based models struggle to identify the sought target in such cases, often leading to tracking failure. Moreover, the target object may undergo a drastic appearance change over time, further complicating the discrimination between target and distractor objects. In certain scenarios, \eg, as visualized in Fig.~\ref{fig:teaser}, it is even virtually impossible to unambiguously identify the target solely based on appearance information. Such circumstances can only be addressed by leveraging other cues during tracking, for instance the spatial relations between objects. We therefore set out to address problematic distractors by exploring such alternative cues.

\input{figures/teaser}

We propose to actively keep track of distractor objects in order to ensure more robust target identification. To this end, we introduce a target candidate association network, that matches distractor objects as well as the target across frames.
Our approach consists of a base appearance tracker, from which we extract target candidates in each frame. Each candidate is encoded with a set of distinctive features, consisting of the target classifier score, location, and appearance.
The encodings of all candidates are jointly processed by a graph-based candidate embedding network.
From the resulting embeddings, we compute the association scores between all candidates in subsequent frames, allowing us to keep track of the target and distractor objects over time. In addition, we estimate a target detection confidence, used to increase the robustness of the target classifier. 

While associating target candidates over time provides a powerful cue, learning such a matching network requires tackling a few key challenges. In particular, generic visual object tracking datasets only provide annotations of one object in each frame, \ie, the target. As a result, there is a lack of ground-truth annotations for associating distractors across frames. 
Moreover, the definition of a distractor is not universal and depends on the properties of the employed appearance model. We address these challenges by introducing an approach that allows our candidate matching network to learn from real tracker output. Our approach exploits the single target annotations in existing tracking datasets in combination with a self-supervised strategy. Furthermore, we actively mine the training dataset in order to retrieve rare and challenging cases, where the use of distractor association is important, in order to learn a more effective model. 

\parsection{Contributions} In summary, our contributions are as follows:
\textbf{(i)} We propose a method for target candidate association based on a learnable candidate matching network.
\textbf{(ii)} We develop an online object association method in order to propagate distractors and the target over time and introduce a sample confidence score to update the target classifier more effectively during inference.
\textbf{(iii)} We tackle the challenges with incomplete annotation by employing partial supervision, self-supervised learning, and sample-mining to train our association network.
\textbf{(iv)} We perform comprehensive experiments and ablative analyses by integrating our approach into the tracker  SuperDiMP~\cite{Danelljan_2019_github_pytracking,Danelljan_2020_CVPR_PRDIMP,Bhat_2019_ICCV_DIMP}. The resulting tracker \emph{KeepTrack} sets a new state-of-the-art on six tracking datasets, obtaining an AUC of $67.1\%$ on LaSOT~\cite{Fan_2019_CVPR_Lasot} and $69.7\%$ on UAV123~\cite{Mueller_2016_ECCV_UAV123}.

%% file: figures/teaser.tex
\begin{figure}[t]
\centering
\includegraphics[width=\columnwidth, keepaspectratio]{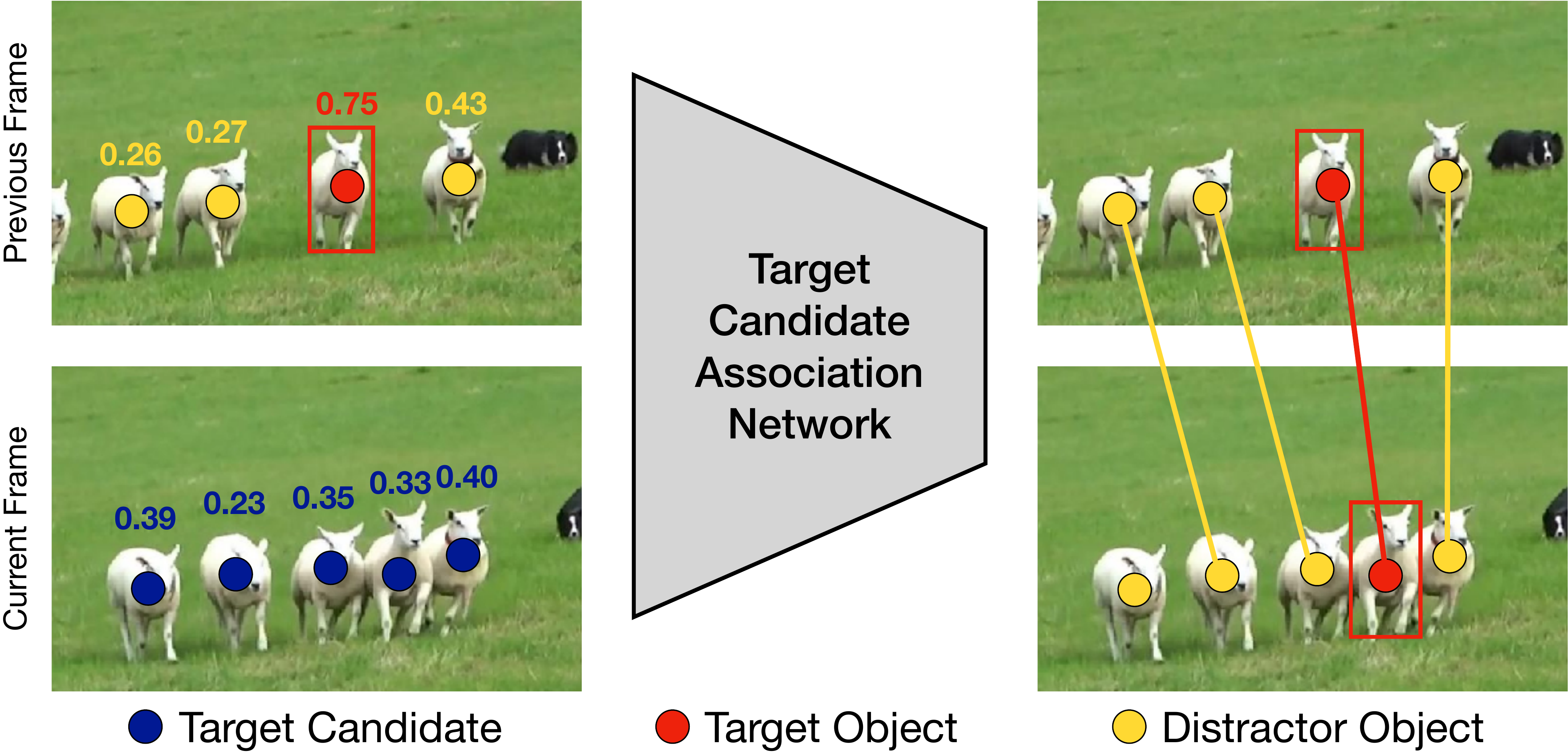}
\caption{Visualization of the proposed target candidate association network used for tracking. For each target candidate (\object{_darkblue}) we extract a set of features such as score, position and appearance in order to associate candidates across frames. The proposed target association network then allows to associate these candidates (\object{_darkblue}) with the detected distractors (\object{_yellow}) and the target object (\object{_red}) of the previous frame. Lines connecting circles represent associations.}\label{fig:teaser}
\end{figure}

%% file: sections/relatedwork.tex
\section{Related Work}

Discriminative appearance model based trackers~\cite{Danelljan_2019_CVPR_ATOM,Bhat_2019_ICCV_DIMP,Henriques_2015_TPAMI_KCF,Galoogahi_2017_ICCV_BACF,Zhang_2014_ECCV_MEEM,Danelljan_2016_ECCV_CCTO} aim to suppress distractors based on their appearance by integrating background information when learning the target classifier online. While often increasing robustness, the capacity of an online appearance model is still limited.
A few works have therefore developed more dedicated strategies of handling distractors.
Bhat~\etal~\cite{Bhat_2020_ECCV_KYS} combine an appearance based tracker and an RNN to propagate information about the scene across frames. It internally aims to track all regions in the scene by maintaining a learnable state representation. Other methods exploit the existence of distractors explicitly in the method formulation. DaSiamRPN~\cite{Zhu_2018_ECCV_DaSiamRPN} handles distractor objects by subtracting corresponding image features from the target template during online tracking. Xiao~\etal~\cite{Xiao_2016_ECCV_DistractorSupportedST} use the locations of distractors in the scene and employ hand crafted rules to classify image regions into background and target candidates on each frame. SiamRCNN~\cite{Voigtlaender_2020_CVPR_SiamRCNN} associates subsequent detections across frames using a hand-crafted association score to form short tracklets.
In contrast, we introduce a learnable network that explicitly associates target candidates from frame-to-frame.
Zhang~\etal~\cite{Zhang_2021_CVPR_DMTrack} propose a tracker inspired by the multi object tracking (MOT) philosophy of tracking by detection. They use the top-k predicted bounding boxes for each frame and link them between frames by using different features. In contrast, we omit any hand crafted features but fully learn to predict the associations using self-supervision.

Many online trackers~\cite{Danelljan_2019_CVPR_ATOM,Bhat_2019_ICCV_DIMP,Danelljan_2017_CVPR_ECO} employ a memory to store previous predictions to fine-tune the tracker. Typically the oldest sample is replaced in the memory and an age-based weight controls the contribution of each sample when updating the tracker online. 
Danelljan~\etal~\cite{Danelljan_2016_CVPR2016_decontamination} propose to learn the tracking model and the training sample weights jointly. 
LTMU~\cite{Dai_2020_CVPR_LTMU} combines an appearance based tracker with a learned meta-updater. The goal of the meta-updater is to predict whether the employed online tracker is ready to be updated or not.
In contrast, we use a learned target candidate association network to compute a confidence score and combine it with sample age to manage the tracker updates. 

The object association problem naturally arises in MOT. The dominant paradigm in MOT is \emph{tracking-by-detection}~\cite{Braso_2020_CVPR_Neural_Solver,Bergmann_2019_ICCV_Tracktor, Yu_2007_CVPR_MOT,Tang_2017_CVPR_MOT, Zhang_2008_CVPR_MOT}, where tracking is posed as the problem of associating object detections over time. The latter is typically formulated as a graph partitioning problem.
Typically, these methods are non-causal and thus require the detections from all frames in the sequence. Furthermore, MOT typically focuses on a limited set of object classes~\cite{dendorfer_2020_IJCV_motchallenge}, such as pedestrians, where strong object detectors are available. In comparison we aim at tracking different objects in different sequences solely defined by the initial frame. Furthermore, we lack ground truth correspondences of all distractor objects from frame to frame whereas the ground-truth correspondences of different objects in MOT datasets are typically provided~\cite{dendorfer_2020_IJCV_motchallenge}. Most importantly, we aim at associating target candidates that are defined by the tracker itself, while MOT methods associate all detections that correspond to one of the sought objects. 

%% file: sections/method.tex
\section{Method}
\input{figures/TCA_blockdiagram}

In this section, we describe our tracking approach, which actively associates distractor objects and the sought target across multiple frames. 

\subsection{Overview}

An overview of our tracking pipeline is shown in Fig.~\ref{fig:diagram}. 
We use a base tracker with a discriminative appearance model and internal memory. In particular, we adopt the SuperDiMP~\cite{Danelljan_2019_github_pytracking,Gustafsson_2020_BMVC_EBM} tracker, which employs the target classifier in DiMP~\cite{Bhat_2019_ICCV_DIMP} and the probabilistic bounding-box regression from~\cite{Danelljan_2020_CVPR_PRDIMP}, together with improved training settings.

We use the base tracker to predict the target score map~$s$ for the current frame and extract the target candidates $v_i$ by finding locations in $s$ with high target score. Then, we extract a set of features for each candidate. Namely: target classifier score $s_i$, location $\vec{c}_i$ in the image, and an appearance cue $\vec{f}_i$ based on the backbone features of the base tracker. Then, we encode this set of features into a single feature vector $\vec{z}_i$ for each candidate. We feed these representations and the equivalent ones of the previous frame -- already extracted beforehand -- into the candidate embedding network and process them together to obtain the enriched embeddings $\vec{h}_i$ for each candidate.  These feature embeddings are used  to compute the similarity matrix $\matrix{S}$  and to estimate the candidate assignment matrix $\matrix{A}$ between the two consecutive frames using an optimal matching strategy.

Once having the candidate-to-candidate assignment probabilities estimated, we build the set of currently visible objects in the scene $\makeSet{O}$ and associate them to the previously identified objects $\makeSet{O}'$,
\ie, we determine which objects disappeared, newly appeared, or stayed visible and can be associated unambiguously.
We then use this propagation strategy to reason about the target object $\hat{o}$ in the current frame. Additionally, we compute the target detection confidence $\beta$ to manage the memory and control the sample weight, while updating the target classifier online.    

\subsection{Problem Formulation}

Let the set of \emph{target candidates}, which includes distractors and the sought target, be $\makeSet{V} = \{v_i\}^N_{i=1}$, where $N$ denotes the number of candidates present in each frame. We define the target candidate sets $\makeSet{V}'$ and $\makeSet{V}$ corresponding to the previous and current frames, respectively. We formulate the problem of target candidate association across two subsequent frames as, finding the assignment matrix $\matrix{A}$ between the two sets $\makeSet{V}'$ and $\makeSet{V}$. If the target candidate $v_{i}'$ corresponds to $v_{j}$ then $\matrix{A}_{i,j} = 1$ and  $\matrix{A}_{i,j} = 0$ otherwise.  

In practice, a match may not exist for every candidate. Therefore, we introduce the concept of dustbins, which is commonly used for graph matching~\cite{Sarlin_2020_CVPR_SuperGlue,DeTone_2018_CVPR_Workshops_SuperPoint} to actively handle the non-matching vertices. The idea is to match the candidates without match to the dustbin on the missing side. Therefore, we augment the assignment matrix $\matrix{A}$ by an additional row and column representing dustbins. It follows that a newly appearing candidate $v_j$ -- which is only present in the set $\makeSet{V}$ -- leads to the entry $\matrix{A}_{N'+1, j} = 1$. Similarly, a candidate $v_{i}'$ that is no longer available in the set $\makeSet{V}$ results in $\matrix{A}_{i, N+1} = 1$. To solve the assignment problem, we design a learnable approach that predicts the matrix $\matrix{A}$. Our approach first extracts a representation of the target candidates, which is discussed below. 

\subsection{Target Candidate Extraction}
Here, we describe how to detect and represent target candidates and propose a set of features and their encoding. We define the set of target candidates $\makeSet{V}$ as all unique coordinates $\vec{c}_i$ that correspond to a local maximum with minimal score in the target score map $s$. Thus, each target candidate $v_i$ and its coordinate $\vec{c}_i$ need to fulfill the following two constraints,
\begin{equation}
    \phi_\mathrm{max}(s, \vec{c}_i) = 1 \quad \mathrm{and}\quad s(\vec{c}_i) \geq \tau, 
\end{equation}
where $\phi_\mathrm{max}$ returns 1 if the score at $\vec{c}_i$ is a local maximum of $s$ or 0 otherwise, and $\tau$ denotes a threshold. This definition allows us to build the sets $\makeSet{V}'$ and $\makeSet{V}$, by retrieving the local maxima of $s'$ and $s$ with sufficient score value. We use the max-pooling operation in a $5\times 5$ local neighbourhood to retrieve the local maxima of $s$ and set $\tau=0.05$.

For each candidate we build a set of features inspired by two observations. First, we notice that the motion of the same objects from frame to frame is typically small and thus similar locations and similar distances between different objects. Therefore, the position $\vec{c}_i$ of a target candidate forms a strong cue. In addition, we observe only small changes in appearance for each object. Therefore, we use the target classifier score $s_i = s(\vec{c}_i)$ as another cue. In order to add a more discriminative appearance based feature $\vec{f}_i = \vec{f}(\vec{c}_i)$, we process the backbone features (used in the baseline tracker) with a single learnable convolution layer.
Finally, we build a feature tuple for each target candidate as $(s_i, \vec{f}_i, \vec{c}_i)$. These features are combined in the following way,
\begin{equation*}
    \vec{z}_i = \vec{f}_i + \psi(s_i, \vec{c}_i), \quad \forall v_i \in \makeSet{V},
\end{equation*}
where $\psi$ denotes a Multi-Layer Perceptron (MLP), that maps $s$ and $\vec{c}$ to the same dimensional space as $\vec{f}_i$.
This encoding permits jointly reasoning about appearance, target similarity, and position.

\subsection{Candidate Embedding Network}

In order to further enrich the encoded features and in particular to facilitate extracting features while being aware of neighbouring candidates, we employ a candidate embedding network.
On an abstract level, our association problem bares similarities with the task of sparse feature matching.
In order to incorporate information of neighbouring candidates, we thus take inspiration from recent advances in this area. In particular, we adopt the SuperGlue~\cite{Sarlin_2020_CVPR_SuperGlue} architecture that establishes the current state-of-the-art in sparse feature matching. Its design allows to exchange information between different nodes, to handle occlusions, and to estimate the assignment of nodes across two images. In particular, the features of both frames translate to nodes of a single complete graph with two types of directed edges: 1) self edges within the same frame and 2) cross edges connecting only nodes between the frames. The idea is then to exchange information either along self or cross edges. 

The adopted architecture~\cite{Sarlin_2020_CVPR_SuperGlue} uses a Graph Neural Network (GNN) with message passing that sends the messages in an alternating fashion across self or cross edges to produce a new feature representation for each node after every layer. Moreover, an attention mechanism computes the messages using self attention for self edges and cross attention for cross edges. After the last message passing layer a linear projection layer extracts the final feature representation $\vec{h}_i$ for each candidate $v_i$. 

\subsection{Candidate Matching}\label{sec:candidate_matching}
To represent the similarities between candidates $v'_i\in \makeSet{V}'$ and $v_j\in \makeSet{V}$, we construct the similarity matrix $\matrix{S}$. 
The sought similarity is measured using the scalar product: $\matrix{S}_{i,j} = \langle \vec{h}'_i, \vec{h}_j \rangle$, for feature vectors $\vec{h}'_i$ and $\vec{h}_j$ corresponding to the candidates $v'_i$ and $v_j$.

As previously introduced, we make use of the dustbin-concept~\cite{DeTone_2018_CVPR_Workshops_SuperPoint,Sarlin_2020_CVPR_SuperGlue} to actively match candidates that miss their counterparts to the so-called dustbin. However, a dustbin is a virtual candidate without any feature representation $\vec{h}_i$. Thus, the similarity score is not directly predictable between candidates and the dustbin. A candidate corresponds to the dustbin, only if its  similarity scores to all other candidates are sufficiently low. 
In this process, the similarity matrix $\matrix{S}$ represents only an initial association prediction between candidates disregarding the dustbins.
Note that a candidate corresponds either to an other candidate or to the dustbin in the other frame. When the candidates $v'_i$ and $v_j$ are matched, both constraints $\sum_{i=1}^{N'} \matrix{A}_{i,j} = 1$ and $\sum_{j=1}^{N} \matrix{A}_{i,j} = 1$  must be satisfied for one-to-one matching. These constraints however, do not apply for missing matches since multiple candidates may correspond to the same dustbin. 
Therefore, we make use of two new constraints for dustbins. These constraints for dustbins read as follows: all candidates not matched to another candidate must be matched to the dustbins. Mathematically, this can be expressed as, 
$\sum_j \matrix{A}_{N'+1,j} = N - M$ and $\sum_i \matrix{A}_{i, N+1}  = N' - M$, where $M = \sum_{(i\leq N', j \leq N)} \matrix{A}_{i,j}$ represents the number of candidate-to-candidate matching. In order to solve the association problem, using the discussed constraints, we follow Sarlin~\etal~\cite{Sarlin_2020_CVPR_SuperGlue} and use the Sinkhorn~\cite{Sinkhorn_1967_PJM_Sinkhorn, Cuturi_2013_NIPS_Sinkhorn} based algorithm therein.  

\subsection{Learning Candidate Association}

Training the embedding network that parameterizes the similarity matrix used for optimal matching requires ground truth assignments.
Hence, in the domain of sparse keypoint matching, recent learning based approaches leverage large scale datasets~\cite{Dusmanu_2019_CVPR_D2Net,Sarlin_2020_CVPR_SuperGlue} such as MegaDepth~\cite{Li_2018_CVPR_MegaDepth} or ScanNet~\cite{Dai_2017_CVPR_ScanNet}, that provide ground truth matches. 
However, in tracking such ground truth correspondences (between distractor objects) are not available.
Only the target object and its location provide a single ground truth correspondence. 
Manually annotating correspondences for distracting candidates, identified by a tracker on video datasets, is expensive and may not be very useful.
Instead, we propose a novel training approach that exploits, (i) partial supervision from the annotated target objects, and (ii) self-supervision by artificially mimicking the association problem. Our approach requires only the annotations that already exist in standard tracking datasets.
The candidates for matching are obtained by running the base tracker on the given training dataset.

\parsection{Partially Supervised Loss}
For each pair of consecutive frames, we retrieve the two candidates corresponding to the annotated target, if available. This correspondence forms a partial supervision for a single correspondence while all other associations remain unknown. For the retrieved candidates $v'_i$ and $v_j$, we define the association as a tuple  $(l',l) = (i, j)$. Here, we also mimic the association for redetections and occlusions by occasionally excluding one of the corresponding candidates from $\makeSet{V}'$ or $\makeSet{V}$. We replace the excluded candidate by the corresponding dustbin to form the correct association for supervision. More precisely, the simulated associations for redetection and occlusion are expressed as, $(l', l) = (N'+1, j)$ and $(l', l) = (i, N+1)$, respectively.  
The supervised loss, for each frame-pairs, is then given by the negative log-likelihood of the assignment probability,
\begin{equation}
    L_\mathrm{sup} = -\log \matrix{A}_{l', l}.
\end{equation}
\parsection{Self-Supervised Loss}
To facilitate the association of distractor candidates, we employ a self-supervision strategy.
The proposed approach first extracts a set of candidates $\makeSet{V}'$ from any given frame. 
The corresponding candidates for matching, say $\makeSet{V}$, are identical to $\makeSet{V}'$ but we augment its features.
Since the feature augmentation does not affect the associations, the initial ground-truth association set is given by $\makeSet{C} = \{(i,i)\}_{i=1}^N$.
In order to create a more challenging learning problem, 
we simulate occlusions and redetections as described above for the partially supervised loss. Note that the simulated occlusions and redetections change the entries of $\makeSet{V}$, $\makeSet{V}'$, and $\makeSet{C}$. We make use of the same notations with slight abuse for simplicity. Our feature augmentation involves,  
randomly translating the location $\vec{c}_i$, increasing or decreasing the score $s_i$, and transforming the given image before extracting the visual features $\vec{f}_i$. Now, using the simulated ground-truth associations $\makeSet{C}$, our self-supervised loss is given by,
\begin{equation}
    L_\mathrm{self} = \sum_{(l',l)\in\makeSet{C}} -\log \matrix{A}_{l', l}.
\end{equation}
Finally, we combine both losses as $L_\mathrm{tot} = L_\mathrm{sup} + L_\mathrm{self}$. 
It is important to note that the real training data is  used only for the former loss function, whereas synthetic data is used only for the latter one. 

\parsection{Data Mining}
Most frames contain a candidate corresponding to the target object and are thus applicable for supervised training. However, a majority of these frames are not very informative for training because they contain only a single candidate with high target classifier score and correspond to the target object. 
Conversely, the dataset contains adverse situations where associating the candidate corresponding to the target object is very challenging. Such situations include sub-sequences with different number of candidates, with changes in appearance or large motion between frames.
Thus, sub-sequences where the appearance model either fails and starts to track a distractor or when the tracker is no longer able to detect the target with sufficient confidence are valuable for training. However, such failure cases are rare even in large scale datasets. Similarly, we prefer frames with many target candidates when creating synthetic sub-sequences to simultaneously include candidate associations, redetections and occlusions. Thus, we mine the training dataset using the dumped predictions of the base tracker to use more informative training samples. 

\parsection{Training Details}
We first retrain the base tracker SuperDiMP without the learned discriminative loss parameters but keep everything else unchanged.  We split the LaSOT training set into a \emph{train-train} and a \emph{train-val} set. We run the base tracker on all sequences and save the search region and score map for each frame on disk. We use the dumped data to mine the dataset and to extract the target candidates and its features.
We freeze the weights of the base tracker during training of the proposed network and train for 15 epochs by sampling 6400 sub-sequences per epoch from the \emph{train-train} split. We sample real or synthetic data equally likely. We use ADAM~\cite{Kingma_2014_ICLR_ADAM} with learning rate decay of 0.2 every 6\emph{th} epoch with a learning rate of 0.0001. We use two GNN Layers and run 10 Sinkhorn iterations.
Please refer to the supplementary (Sec.~\ref{sup:sec:training}) for additional details about training.

\subsection{Object Association}
\label{sec:objec_association}

This part focuses on using the estimated assignments (see Sec.~\ref{sec:candidate_matching}) in order to determine the object correspondences during online tracking. An object corresponds either to the target or a distractor.
The general idea is to keep track of every object present in each scene over time. We implement this idea with a database $\makeSet{O}$, where each entry corresponds to an object $o$ that is visible in the current frame.
Fig.~\ref{fig:sequence_comparison} shows these objects as circles. 
An object disappears from the scene if none of the current candidates is associated with it, \eg, in Fig.~\ref{fig:sequence_comparison} the purple and pink objects (\object{_purple},~\object{_pink}) no longer correspond to a candidate in the last frame. Then, we delete this object from the database. Similarly, we add a new object to the database if a new target candidate appears (\object{_orange}, \object{_blue}, \object{_pink}). When initializing a new object, we assign it a new object-id (not used previously) and the score $s_i$. In Fig.~\ref{fig:sequence_comparison} object-ids are represented using colors. For objects that remain visible, we add the score $s_i$ of the corresponding candidate to the history of scores of this object. 
Furthermore, we delete the old and create a new object if the candidate correspondence is ambiguous, \ie, the assignment probability is smaller than $\omega = 0.75$.

If associating the target object $\hat{o}$ across frames is unambiguous, it implies that one object has the same object-id as the initially provided object $\hat{o}_\mathrm{init}$. Thus, we return this object as the selected target. 
However, in real world scenarios the target object gets occluded, leaves the scene or associating the target object is ambiguous. Then, none of the candidates corresponds to the sought target and we  need to redetect.
We redetect the target if the candidate with the highest target classifier score achieves a score that exceeds the threshold $\eta = 0.25$.
We select the corresponding object as the target as long as no other candidate achieves a higher score in the current frame. Then, we switch to this candidate and declare it as target if its score is higher than any score in the history (of the currently selected object). Otherwise, we treat this object as a distractor for now, but if its score increases high enough, we will select it as the target object in the future.
Please refer to the supplementary material (Sec.~\ref{sup:sec:inference}) for the detailed algorithm. 

\input{figures/sequence_comparison}

\subsection{Memory Sample Confidence}
\label{sec:memory}
While updating the tracker online is often beneficial, it is disadvantageous if the training samples have a poor quality. Thus, we describe a memory sample confidence score, that we use to decide which sample to keep in the memory and which should be replaced when employing a fixed size memory. In addition, we use the score to control the contribution of each training sample when updating the tracker online. In contrast, the base tracker replaces frames using a first-in-first out policy if the target was detected and weights samples during inference solely based on age.

First, we define the training samples in frame $k$ as $(x_k,y_k)$. We assume a memory size $m$ that stores samples from frame $k\in\{1,\dots,t\}$, where $t$ denotes the current frame number. The  online loss then given by,
\begin{equation}\label{eq:online-loss}
    J(\theta) = \lambda R(\theta) + \sum_{k=1}^t \alpha_k\beta_k Q(\theta; x_k, y_k) ,
\end{equation}
where $Q$ denotes the data term, $R$ the regularisation term, $\lambda$ is a scalar and $\theta$ represents appearance model weights.
The weights $\alpha_k\geq0$ control the impact of the sample from frame $k$, \ie, a higher value increases the influence of the corresponding sample during training. We follow other appearance based trackers~\cite{Bhat_2019_ICCV_DIMP,Danelljan_2019_CVPR_ATOM} and use a learning parameter $\gamma\in[0,1]$ in order to control the weights $\alpha_k = (1-\gamma)\alpha_{k+1}$, such that older samples achieve a smaller value and their impact during training decreases. In addition, we propose a second set of weights $\beta_k$ that represent the confidence of the tracker that the predicted label $y_k$ is correct.
Instead of removing the oldest samples to keep the memory fixed~\cite{Bhat_2019_ICCV_DIMP}, we propose to drop the sample that achieves the smallest score $\alpha_k\beta_k$ which combines age and confidence.  Thus, if $t > n$ we remove the sample at position $k = \mathrm{argmin}_{1\leq k\leq n} \alpha_k\beta_k$ by setting $\alpha_k = 0$. This means, that if all samples achieve similar confidence the oldest is replaced, or that if all samples are of similar age the least confident sample is replaced.

We describe the extraction of the confidence weights as,
\begin{equation}\label{confidence_reweighting}
    \beta_t=
    \begin{cases}
        \sqrt{\sigma}, & \text{if}\ \hat{o} = \hat{o}_\mathrm{init} \\
        \sigma, & \text{otherwise},
    \end{cases}
\end{equation}
where $\sigma = \max_i s_i^t$ denotes the maximum value of the target classifier score map of frame $t$. For simplicity, we assume that $\sigma \in [0, 1]$. The condition $\hat{o} = \hat{o}_\mathrm{init}$ is fulfilled if the currently selected object is identical to the initially provided target object, \ie, both objects share the same object id. Then, it is very likely, that the selected object corresponds to the
target object such that we increase the confidence using the square root function that increases values in the range $[0, 1)$.
Hence, the described confidence score combines the confidence of the target classifier with the confidence of the object association module, but fully relies on the target classifier once the target is lost.

\parsection{Inference details}
We propose \emph{KeepTrack} and the speed optimized \emph{KeepTrackFast}. We use the SuperDiMP parameters for both trackers but increase the search area scale from 6 to 8 (from 352 to 480 in image space) for \emph{KeepTrack}. For the fast version we keep the original scale but reduce the number of bounding box refinement steps from 10 to 3.
In addition, we skip running the association module if only one target caidndate with a high score is present in the previous and current frame.
Overall, both trackers follow the target longer until it is lost such that small search areas occur frequently. Thus, we reset the search area to its previous size if it was drastically decreased before the target was lost, to facilitate redetections.
Please refer to the supplementary (Sec.~\ref{sup:sec:inference}) for more details.

%% file: figures/TCA_blockdiagram.tex
\begin{figure*}[t]
\centering
\includegraphics[width=\textwidth, keepaspectratio]{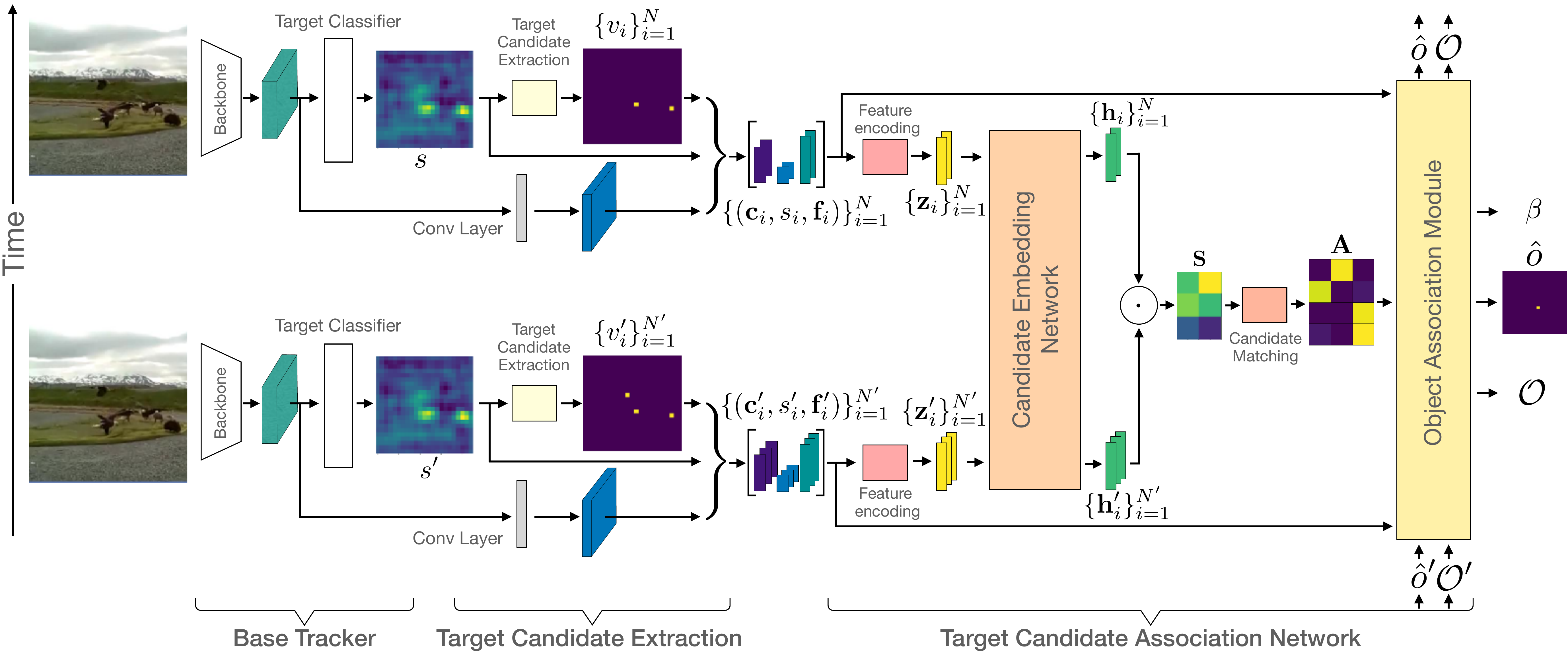}
\caption{Overview of the entire online tracking pipeline, processing the previous and current frames jointly to predict the target object.
}\label{fig:diagram}
\end{figure*}

%% file: figures/sequence_comparison.tex
\begin{figure}[t]
\centering
\includegraphics[width=\columnwidth, keepaspectratio]{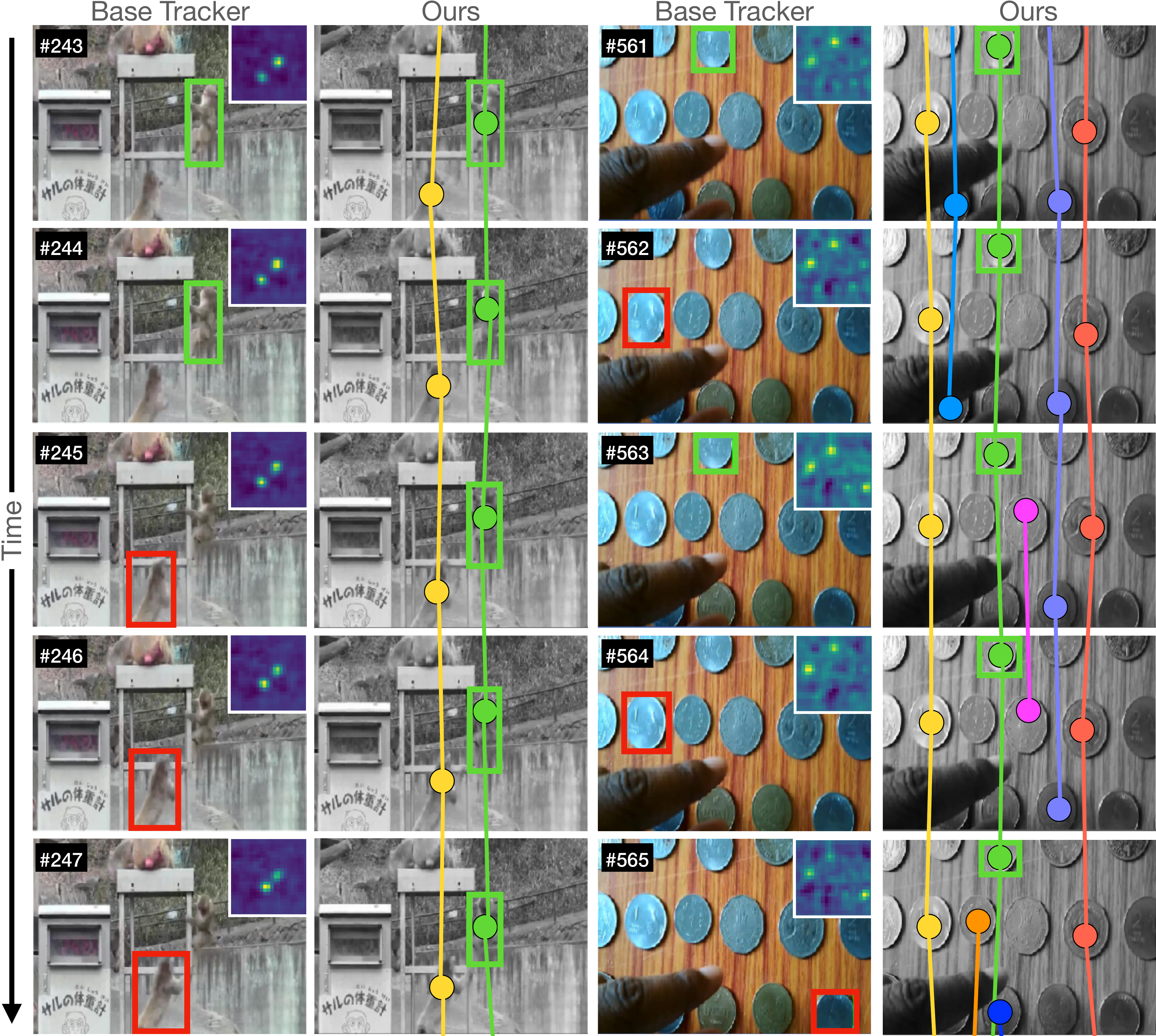}
\caption{Visual comparison of the base tracker and our tracker. The bounding boxes represent the tracker result, green [\bbox{_green}] indicates correct detections and red [\bbox{_red}] refers to tracker failure. Each circle represents an object. Circles with the same color are connected to indicate that the object-ids are identical. If a target candidate cannot be matched with an existing object we add a new object (\object{_orange}, \object{_blue}, \object{_pink}). Similarly, we delete the object if no candidate corresponds to it anymore in the next frame (\object{_skyblue}, \object{_purple}, \object{_pink}).
}\label{fig:sequence_comparison}
\vspace{-3mm}
\end{figure}

%% file: sections/experiments_and_results.tex
\section{Experiments}

We evaluate our proposed tracking architecture on seven benchmarks. Our approach is implemented in Python using PyTorch.
On a single Nvidia GTX 2080Ti GPU, \emph{KeepTrack} and \emph{KeepTrackFast} achieve 18.3 and 29.6~FPS, respectively.

\subsection{Ablation Study}
\label{sec:ablation_study}
We perform an extensive analysis of the proposed tracker, memory sample confidence, and training losses.

\parsection{Online tracking components}
\input{tables/components_ablation}
We study the importance of memory sample confidence, the search area protocol, and target candidate association of our final method \emph{KeepTrack}. In Tab.~\ref{tab:component_ablation} we analyze the impact of successively adding each component, and report the average of five runs on the NFS~\cite{Galoogahi_2017_ICCV_NFS}, UAV123~\cite{Mueller_2016_ECCV_UAV123} and LaSOT~\cite{Fan_2019_CVPR_Lasot} datasets. The first row reports the results of the employed base tracker. First, we add the memory sample confidence approach (Sec.~\ref{sec:memory}), observe similar performance on NFS and UAV but a significant improvement of $1.5\%$ on LaSOT, demonstrating its potential for long-term tracking. 
With the added robustness, we next employ a larger search area and increase it if it was drastically shrank before the target was lost. This leads to a fair improvement on all datasets.
Finally, we add the target candidate association network, which provides substantial performance improvements on all three datasets, with a $+1.3\%$ AUC on LaSOT. These results clearly demonstrate the power of the target candidate association network.

\parsection{Training}
\input{tables/training_loss_ablation}
\input{tables/mem_components_ablation}
In order to study the effect of the proposed training losses, we retrain the target candidate association network either with only the partially supervised loss or only the self-supervised loss. We report the performance on LaSOT~\cite{Fan_2019_CVPR_Lasot} in Tab.~\ref{tab:training_loss_ablation}. The results show that each loss individually allows to train the network and to outperform the baseline without the target candidate association network (no TCA), whereas, combining both losses leads to the best tracking results. In addition, training the network with the combined loss but without data-mining decreases the tracking performance.

\parsection{Memory management}
We not only use the sample confidence to manage the memory but also to control the impact of samples when learning the target classifier online. In Tab.~\ref{tab:mem_component_ablation}, we study the importance of each component by adding one after the other and report the results on LaSOT~\cite{Fan_2019_CVPR_Lasot}. First, we use the sample confidence scores only to decide which sample to remove next from the memory. This, already improves the tracking performance. Reusing these weights when learning the target classifier as described in Eq.~\eqref{eq:online-loss} increases the performance again. To suppress the impact of poor-quality samples during online learning, we ignore samples with a confidence score bellow $0.5$. This leads to an improvement on LaSOT. The last row corresponds to the used setting in the final proposed tracker.

\subsection{State-of-the-art Comparison}\label{subsec:experiments}
We compare our approach on seven tracking benchmarks. 
The same settings and parameters are used for all datasets.
In order to ensure the significance of the results, we report the average over five runs on all datasets unless the evaluation protocol requires otherwise. 
We recompute the results of all trackers using the raw predictions if available or otherwise report the results given in the paper.

\input{tables/lasot}

\parsection{LaSOT~\cite{Fan_2019_CVPR_Lasot}}
First, we compare on the large-scale LaSOT dataset (280 test sequences with 2500 frames in average) to demonstrate the robustness and accuracy of the proposed tracker. The success plot in Fig.~\ref{fig:lasot} shows the overlap precision $\text{OP}_T$ as a function of the threshold $T$. 
Trackers are ranked \wrt their \emph{area-under-the-curve} (AUC) score, denoted in the legend. Tab.~\ref{tab:lasot} shows more results including precision and normalized precision. \emph{KeepTrack} and \emph{KeepTrackFast} outperform the recent trackers AlphaRefine~\cite{Yan_2021_CVPR_AlphaRefine}, TransT~\cite{Chen_2021_CVPR_TransT} and TrDiMP~\cite{Wang_2021_CVPR_TrDiMP} by a large margin and the base tracker SuperDiMP by $4.0\%$ or $3.7\%$ in AUC. The improvement in $\text{OP}_T$ is most prominent for thresholds $T<0.7$, demonstrating the superior robustness of our tracker. In Tab.~\ref{tab:lasot_super_dimp_base_tracker}, we further perform an apple-to-apple comparison with KYS~\cite{Bhat_2020_ECCV_KYS}, LTMU~\cite{Dai_2020_CVPR_LTMU}, AlphaRefine~\cite{Yan_2021_CVPR_AlphaRefine} and TrDiMP~\cite{Wang_2021_CVPR_TrDiMP}, where all methods use SuperDiMP as base tracker. We outperform the best method on each metric, achieving an AUC improvement of $1.8\%$.
\input{figures/success_plots}
\input{tables/lasot_super_dimp_base_tracker}

\parsection{LaSOTExtSub~\cite{Fan_2020_IJCV_Lasot_ext}}
We evaluate our tracker on the recently published extension subset of LaSOT. LaSOTExtSub is meant for testing only and consists of 15 new classes with 10 sequences each. The sequences are long (2500 frames on average), rendering substantial challenges. Fig.~\ref{fig:lasot_ext_sub} shows the success plot, that is averaged over 5 runs. All results, except ours and SuperDiMP, are obtained from~\cite{Fan_2020_IJCV_Lasot_ext}, \eg, DaSiamRPN~\cite{Zhu_2018_ECCV_DaSiamRPN}, SiamRPN++~\cite{Li_2019_CVPR_SiamRPN++} and SiamMask~\cite{Wang_2019_CVPR_SiamMask}. Our method achieves superior results, outperforming LTMU by $6.8\%$ and SuperDiMP by $3.5\%$. 

\parsection{OxUvALT~\cite{Valmadre_2018_ECCV_OxUvA}}
The OxUvA long-term dataset contains 166 test videos with average length 3300 frames. Trackers are required to predict whether the target is present or absent in addition to the bounding box for each frame. Trackers are ranked by the maximum geometric mean (MaxGM) of the true positive (TPR) and true negative rate (TNR). We use the proposed confidence score and set the threshold for target presence using the separate dev.\ set.
Tab.~\ref{tab:oxuva} shows the results on the test set, which are obtained through the evaluation server. \emph{KeepTrack} sets the new state-of-the-art in terms of MaxGM by achieving an improvement of $5.8\%$ compared to the previous best method and exceed the result of the base tracker SuperDiMP by $6.1\%$. 

\parsection{VOT2019LT~\cite{Kristan_2019_ICCVW_VOT2019}/VOT2020LT~\cite{Kristan_2020_ECCVW_VOT2020}}
The dataset for both VOT~\cite{Kristan_2016_TPAMI_VOT} long-term tracking challenges  contains 215,294 frames divided in 50 sequences. Trackers need to predict a confidence score that the target is present and the bounding box for each frame. Trackers are ranked by the F-score, evaluated for a range of confidence thresholds. We compare with the top methods in the challenge~\cite{Kristan_2019_ICCVW_VOT2019,Kristan_2020_ECCVW_VOT2020}, as well as more recent methods. As shown in Tab.~\ref{tab:vot2019lt}, our tracker achieves the best result in terms of F-score and outperforms the base tracker SuperDiMP by $4.0\%$ in F-score.

\parsection{UAV123~\cite{Mueller_2016_ECCV_UAV123}}
This dataset contains 123 videos and is designed to assess trackers for UAV applications. It contains small objects, fast motions, and distractor objects.
Tab.~\ref{tab:nfs_uav_otb} shows the results, where the entries correspond to AUC for $\text{OP}_T$ over IoU thresholds $T$. Our method sets a new state-of-the-art with an AUC of $69.7\%$, exceeding the performance of the recent trackers TransT~\cite{Chen_2021_CVPR_TransT} and TrDiMP~\cite{Wang_2021_CVPR_TrDiMP} by 0.6\% and 2.2\% in AUC.

\parsection{OTB-100~\cite{WU_2015_TPAMI_OTB}}
For reference, we also evaluate our tracker on the OTB-100 dataset consisting of 100 sequences. Several trackers achieve tracking results over 70\% in AUC, as shown in Tab.~\ref{tab:nfs_uav_otb}. So do \emph{KeepTrack} and \emph{KeepTrackFast} that perform similarly to the top methods, with a $0.8\%$ and $1.1\%$ AUC gain over the SuperDiMP baseline.

\parsection{NFS~\cite{Galoogahi_2017_ICCV_NFS}}
Lastly, we report results on the 30 FPS version of the Need for Speed (NFS) dataset. It contains fast motions and challenging distractors. Tab.~\ref{tab:nfs_uav_otb} shows that our approach sets a new state-of-the-art on NFS.

\input{tables/oxuva}

\input{tables/vot2019lt}

\input{tables/otb_nfs_uav}

%% file: tables/components_ablation.tex
\begin{table}[!b]
	\centering
	\newcommand{\best}[1]{\textbf{#1}}
	\newcommand{\dist}{\hspace{5pt}}%
	\newcommand{\yes}{\textcolor{black}{\checkmark}}
	\resizebox{\columnwidth}{!}{%
        \begin{tabular}{c@{\dist}c@{\dist}c@{\dist}|c@{\dist}c@{\dist}c@{\dist}}
        	\toprule
        	         Memory Sample& Search area & Target Candidate    &  \\
        	         Confidence   & Adaptation  & Association Network & NFS & UAV123 & LaSOT  \\
        	\midrule
        	 \yes & --   & --   & 64.7 & 68.0 & 65.0 \\
        	 \yes & \yes & --   & 65.2 & 69.1 & 65.8 \\
        	 \yes & \yes & \yes & \best{66.4} & \best{69.7} & \best{67.1} \\\bottomrule
        \end{tabular}
	}\vspace{1mm}%
	\caption{Impact of each component in terms of AUC (\%) on three datasets. The first row corresponds to our SuperDiMP baseline.
	}\vspace{0mm}
	\label{tab:component_ablation}%
\end{table}

%% file: tables/training_loss_ablation.tex
\begin{table}[!b]
	\centering
	\newcommand{\best}[1]{\textbf{#1}}
	\newcommand{\dist}{\hspace{10pt}}%
	\newcommand{\yes}{\textcolor{black}{\checkmark}}
	\resizebox{0.9\columnwidth}{!}{%
        \begin{tabular}{l@{\dist}c@{\dist}c@{\dist}c@{\dist}c@{\dist}c@{\dist}}
        	\toprule
	        \textbf{Loss} &  no TCA & $L_\mathrm{sup}$ & $L_\mathrm{self}$ & $L_\mathrm{sup} + L_\mathrm{self}$ & $L_\mathrm{sup} + L_\mathrm{self}$ \\ 
	        \textbf{Data-mining} & n.a. & \yes & \yes & - & \yes  \\
        	\midrule
            LaSOT, AUC (\%) & 65.8 & 66.0 & 66.9 & 66.8 & \best{67.1}\\\bottomrule
        \end{tabular}
	}
	\caption{Analysis on LaSOT of association learning using different loss functions with and without data-mining.
	}\vspace{-0mm}
	\label{tab:training_loss_ablation}%
\end{table}

%% file: tables/mem_components_ablation.tex
\begin{table}[!t]
	\centering
	\newcommand{\best}[1]{\textbf{#1}}
	\newcommand{\opt}[1]{\textbf{\textcolor{violet}{#1}}}
	\newcommand{\fast}[1]{\textbf{\textcolor{orange}{#1}}}
	\newcommand{\dist}{\hspace{10pt}}%
	\newcommand{\yes}{\textcolor{black}{\checkmark}}
	\resizebox{0.80\columnwidth}{!}{%
        \begin{tabular}{c@{\dist}c@{\dist}c@{\dist}|c@{\dist}}
        	\toprule
        	         
        	         Sample Replacement   & Online updating      & Conf. score       & LaSOT\\
        	         with conf. score  & with conf. score        & threshold  & AUC (\%)  \\
        	\midrule
             --   & --   & --   &  63.5 \\
        	 \yes & --   & --   &  64.1 \\
        	 \yes & \yes & 0.0 &  64.6 \\
        	 \yes & \yes & 0.5 &  65.0 \\\bottomrule
        \end{tabular}
	}%
	\caption{Analysis of our memory weighting component on LaSOT.}
	\label{tab:mem_component_ablation}%
	\vspace{-4mm}
\end{table}

%% file: tables/lasot.tex
\begin{table}[!b]
	\centering
	\vspace{-3mm}
	\newcommand{\best}[1]{\textbf{\textcolor{red}{#1}}}
	\newcommand{\scnd}[1]{\textbf{\textcolor{blue}{#1}}}
	\newcommand{\opt}[1]{\textbf{\textcolor{violet}{#1}}}
	\newcommand{\fast}[1]{\textbf{\textcolor{orange}{#1}}}
	\newcommand{\dist}{\hspace{4pt}}%
	\resizebox{1.00\columnwidth}{!}{%
        \begin{tabular}{l@{\dist}c@{\dist}c@{\dist}c@{\dist}c@{\dist}c@{\dist}c@{\dist}c@{\dist}c@{\dist}c@{\dist}c@{\dist}c@{\dist}c@{\dist}c@{\dist}c@{\dist}c@{\dist}c@{\dist}c@{\dist}c@{\dist}c@{\dist}c@{\dist}c@{\dist}c@{\dist}c@{\dist}c@{\dist}c@{\dist}c@{\dist}c@{\dist}c@{\dist}c@{\dist}}
        	\toprule
        	               & \textbf{Keep}  & \textbf{Keep}  & Alpha  &        & Siam  &        & Super & STM   & Pr   & DM    &      &      &       \\
        	               & \textbf{Track} & \textbf{Track} & Refine & TransT & R-CNN & TrDiMP & Dimp  & Track & DiMP & Track & LTMU & DiMP & Ocean \\
        	               & & \textbf{Fast} & \cite{Yan_2021_CVPR_AlphaRefine} & \cite{Chen_2021_CVPR_TransT} & \cite{Voigtlaender_2020_CVPR_SiamRCNN} & \cite{Wang_2021_CVPR_TrDiMP} & \cite{Danelljan_2019_github_pytracking} & \cite{Fu_2021_CVPR_STMTrack} & \cite{Danelljan_2020_CVPR_PRDIMP} & \cite{Zhang_2021_CVPR_DMTrack} & \cite{Dai_2020_CVPR_LTMU} & \cite{Bhat_2019_ICCV_DIMP} & \cite{Zhang_2020_ECCV_Ocean}  \\
        	\midrule
        	
        	Precision      & \best{70.2}  & \scnd{70.0} & 68.0 & 69.0 & 68.4 & 66.3 & 65.3 & 63.3 & 60.8 & 59.7 & 57.2 & 56.7 & 56.6 \\
        	Norm. Prec     & \best{77.2}  & \scnd{77.0} & 73.2 & 73.8 & 72.2 & 73.0 & 72.2 & 69.3 & 68.8 & 66.9 & 66.2 & 65.0 & 65.1 \\
        	Success (AUC)  & \best{67.1}  & \scnd{66.8} & 65.3 & 64.9 & 64.8 & 63.9 & 63.1 & 60.6 & 59.8 & 58.4 & 57.2 & 56.9 & 56.0 \\\bottomrule
        \end{tabular}
	}
	\caption{State-of-the-art comparison on the LaSOT~\cite{Fan_2019_CVPR_Lasot} test set in terms of AUC score. 
	}\vspace{-0mm}
	\label{tab:lasot}%
\end{table}

%% file: figures/success_plots.tex
\begin{figure}[!t]
	\centering\vspace{-5mm}%
	\newcommand{\wid}{0.5\columnwidth}%
	\subfloat[LaSOT~\cite{Fan_2019_CVPR_Lasot}\label{fig:lasot}]{\includegraphics[width=\wid]{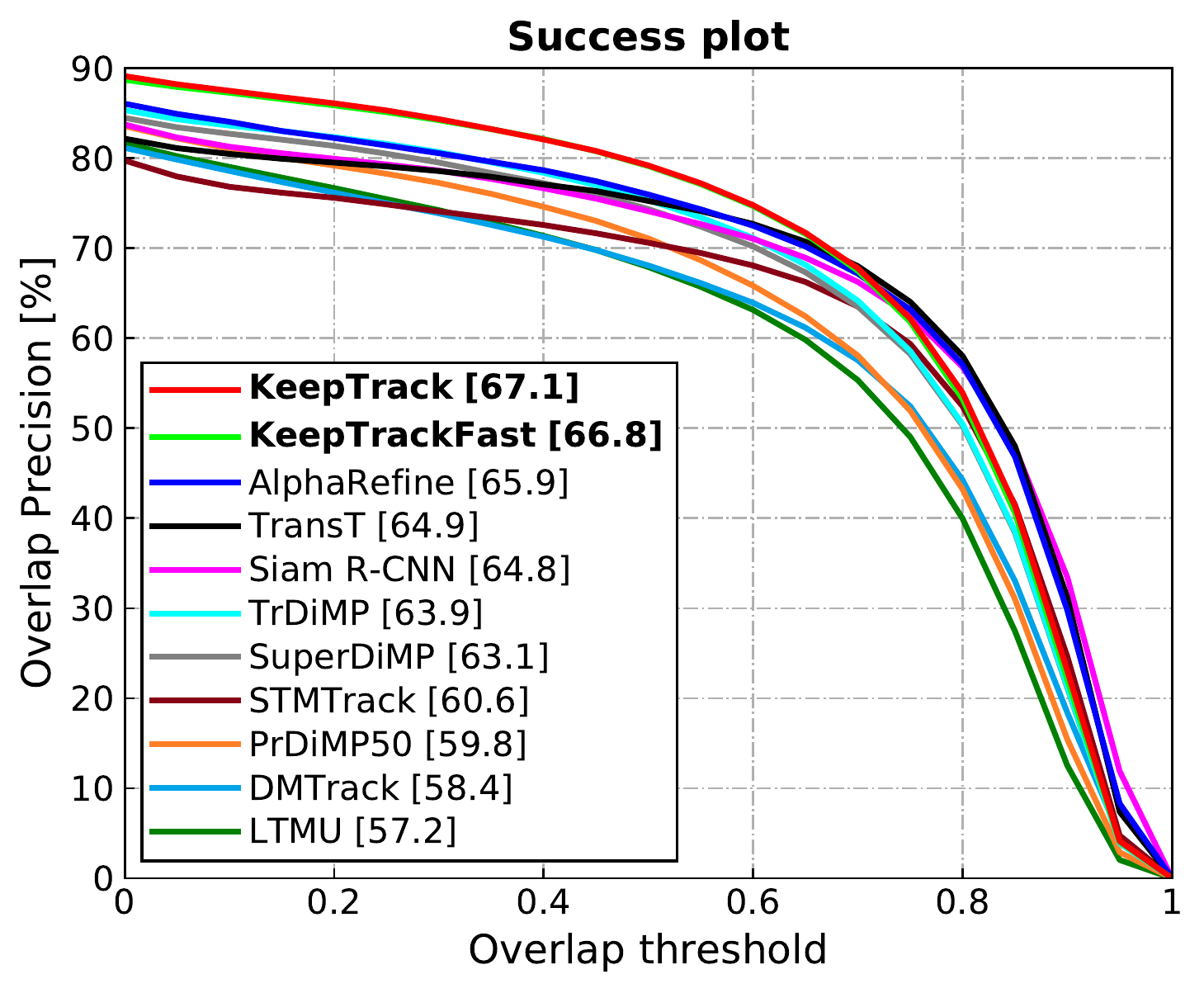}}~%
	\subfloat[LaSOTExtSub~\cite{Fan_2020_IJCV_Lasot_ext}\label{fig:lasot_ext_sub}]{\includegraphics[width = \wid]{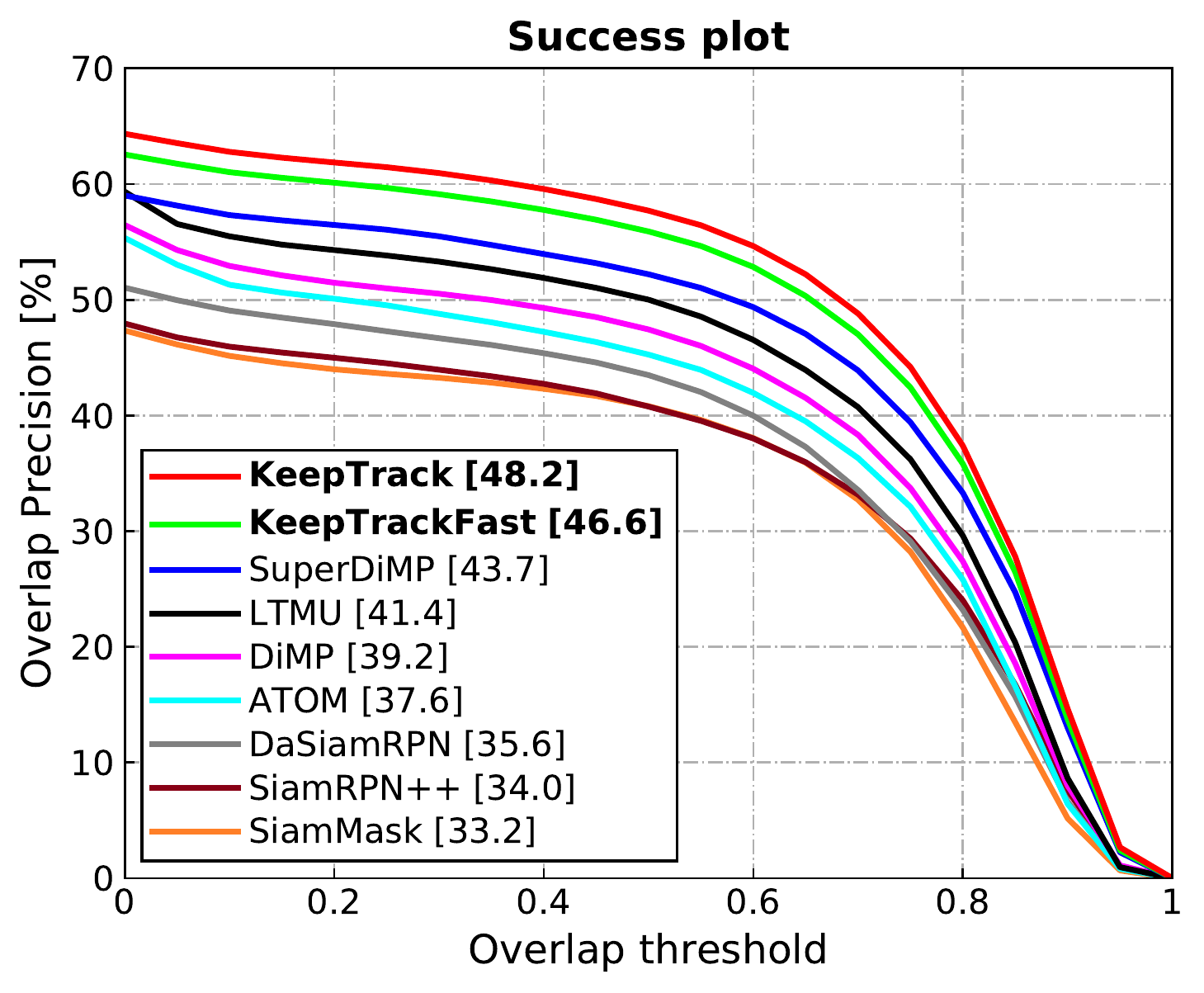}}%
	\vspace{0mm}%
	\caption{Success plots, showing $\text{OP}_T$, on LaSOT~\cite{Fan_2019_CVPR_Lasot} and LaSOTExtSub~\cite{Fan_2020_IJCV_Lasot_ext}. Our approach outperforms all other methods by a large margin in AUC, reported in the legend.}%
	\label{fig:success}\vspace{-3mm}%
\end{figure}

%% file: tables/lasot_super_dimp_base_tracker.tex
\begin{table}[!t]
    \vspace{-0mm}
	\centering
	\newcommand{\best}[1]{\textbf{\textcolor{red}{#1}}}
	\newcommand{\scnd}[1]{\textbf{\textcolor{blue}{#1}}}
	\newcommand{\opt}[1]{\textbf{\textcolor{violet}{#1}}}
	\newcommand{\fast}[1]{\textbf{\textcolor{orange}{#1}}}
	\newcommand{\dist}{\hspace{10pt}}%
	\resizebox{1\columnwidth}{!}{%
        \begin{tabular}{l@{\dist}c@{\dist}c@{\dist}c@{\dist}c@{\dist}c@{\dist}c@{\dist}c@{\dist}}
        	\toprule
        	   & \textbf{KeepTrack} & \textbf{KeepTrack} & AlphaRefine & LTMU & TrDiMP & KYS & SuperDiMP \\
        	   & & \textbf{Fast} & \cite{Yan_2021_CVPR_AlphaRefine} & \cite{Dai_2020_CVPR_LTMU} & \cite{Wang_2021_CVPR_TrDiMP} & \cite{Bhat_2020_ECCV_KYS} & \cite{Danelljan_2020_CVPR_PRDIMP} \\
        	\midrule
        	Precision     & \best{70.2} & \scnd{70.0} & 68.0 & 66.5 & 61.4 & 64.0 & 65.3  \\
        	Norm. Prec.   & \best{77.2} & \scnd{77.0} & 73.2 & 73.7 & --   & 70.7 & 72.2 \\
        	Success (AUC) & \best{67.1} & \scnd{66.8} & 65.3 & 64.7 & 63.9 & 61.9 & 63.1 \\\bottomrule
        \end{tabular}
	}
	\caption{Results on the LaSOT~\cite{Fan_2019_CVPR_Lasot} test set. All trackers use the same base tracker SuperDiMP~\cite{Danelljan_2019_github_pytracking}.
	}\vspace{-3mm}
	\label{tab:lasot_super_dimp_base_tracker}%
\end{table}

%% file: tables/oxuva.tex
\begin{table}[!t]
	\centering
	\newcommand{\best}[1]{\textbf{\textcolor{red}{#1}}}
	\newcommand{\scnd}[1]{\textbf{\textcolor{blue}{#1}}}
	\newcommand{\opt}[1]{\textbf{\textcolor{violet}{#1}}}
	\newcommand{\fast}[1]{\textbf{\textcolor{orange}{#1}}}
	\newcommand{\dist}{\hspace{10pt}}%
	\resizebox{1.00\columnwidth}{!}{%
        \begin{tabular}{l@{\dist}c@{\dist}c@{\dist}c@{\dist}c@{\dist}c@{\dist}c@{\dist}c@{\dist}c@{\dist}c@{\dist}c@{\dist}c@{\dist}c@{\dist}}
        	\toprule
        	   & \textbf{Keep}  & \textbf{Keep}  &      & Super & Siam  &      & DM    &      & Global &      & Siam &     \\
        	   & \textbf{Track} & \textbf{Track} & LTMU & DiMP  & R-CNN & TACT & Track & SPLT & Track  & MBMD & FC+R & TLD \\
        	   &                & \textbf{Fast}  & \cite{Dai_2020_CVPR_LTMU} & \cite{Danelljan_2019_github_pytracking} & \cite{Voigtlaender_2020_CVPR_SiamRCNN} & \cite{Choi_2020_ACCV_TACT} & \cite{Zhang_2021_CVPR_DMTrack} & \cite{Yan_2019_ICCV_SPLT} & \cite{Huang_2020_AAAI_GlobalTrack} & \cite{Zhang_2018_IJCV_MBMD} & \cite{Valmadre_2018_ECCV_OxUvA} & \cite{Kalal_2012_TPAMI_TLD}\\
        	\midrule
        	TPR            & 80.6        & \best{82.7}  & 74.9 & 79.7 & 70.1        & \scnd{80.9} & 68.6 & 49.8 & 57.4 & 60.9 & 42.7     & 20.8 \\
        	TNR            & \scnd{81.2} & 77.2         & 75.4 & 70.2 & 74.5        & 62.2        & 69.4 & 77.6 & 63.3 & 48.5 & 48.1     & \best{89.5} \\
        	\textbf{MaxGM} & \best{80.9} & \scnd{79.9}  & 75.1 & 74.8 & 72.3        & 70.9        & 68.8 & 62.2 & 60.3 & 54.4 & 45.4     & 43.1 \\\bottomrule
        \end{tabular}
	}\vspace{1mm}
	\caption{Results on the OxUvALT~\cite{Valmadre_2018_ECCV_OxUvA} test set in terms of TPR, TNR, and the max geometric mean (MaxGM) of TPR and TNR.
	}\vspace{-3mm}
	\label{tab:oxuva}%
\end{table}

%% file: tables/vot2019lt.tex
\begin{table}[!t]
	\centering
	\newcommand{\best}[1]{\textbf{\textcolor{red}{#1}}}
	\newcommand{\scnd}[1]{\textbf{\textcolor{blue}{#1}}}
	\newcommand{\dist}{\hspace{10pt}}%
	\resizebox{1.00\columnwidth}{!}{%
        \begin{tabular}{l@{\dist}c@{\dist}c@{\dist}c@{\dist}c@{\dist}c@{\dist}c@{\dist}c@{\dist}c@{\dist}c@{\dist}c@{\dist}c@{\dist}}
        	\toprule
        	          & \textbf{Keep}  & \textbf{Keep}  &         &         & Mega  &      & RLT  & Super & Siam   &         \\
        	          & \textbf{Track} & \textbf{Track} & LT\_DSE & LTMU\_B & track & CLGS & DiMP & DiMP  & DW\_LT & ltMBNet \\
        	          &                & \textbf{Fast}  & \cite{Kristan_2019_ICCVW_VOT2019,Kristan_2020_ECCVW_VOT2020} & \cite{Dai_2020_CVPR_LTMU,Kristan_2020_ECCVW_VOT2020} & \cite{Kristan_2020_ECCVW_VOT2020} & \cite{Kristan_2019_ICCVW_VOT2019,Kristan_2020_ECCVW_VOT2020} & \cite{Kristan_2020_ECCVW_VOT2020} & \cite{Danelljan_2019_github_pytracking} & \cite{Kristan_2019_ICCVW_VOT2019,Kristan_2020_ECCVW_VOT2020} & \cite{Kristan_2020_ECCVW_VOT2020} \\
        	\midrule
        	Precision          & \scnd{72.3} & 70.6 & 71.5        & 70.1 & 70.3 & \best{73.9} & 65.7 & 67.6 & 69.7 & 64.9 \\
        	Recall             & \best{69.7} & 68.0 & 67.7        & 68.1 & 67.1 & 61.9        & 68.4 & 66.3 & 63.6 & 51.4 \\
        	\textbf{F-Score}   & \best{70.9} & 69.3 & \scnd{69.5} & 69.1 & 68.7 & 67.4        & 67.0 & 66.9 & 66.5 & 57.4 \\\bottomrule
        \end{tabular}
	}\vspace{1mm}
	\caption{Results on the VOT2019LT~\cite{Kristan_2019_ICCVW_VOT2019}/VOT2020LT~\cite{Kristan_2020_ECCVW_VOT2020} dataset in terms of F-Score, Precision and Recall.}\vspace{-3mm}
	\label{tab:vot2019lt}%
\end{table}

%% file: tables/otb_nfs_uav.tex
\begin{table}[!t]
	\centering
	\newcommand{\best}[1]{\textbf{\textcolor{red}{#1}}}
	\newcommand{\scnd}[1]{\textbf{\textcolor{blue}{#1}}}
	\newcommand{\opt}[1]{\textbf{\textcolor{violet}{#1}}}
	\newcommand{\fast}[1]{\textbf{\textcolor{orange}{#1}}}
	\newcommand{\dist}{\hspace{5pt}}%
	\resizebox{1.00\columnwidth}{!}{%
        \begin{tabular}{l@{\dist}c@{\dist}c@{\dist}c@{\dist}c@{\dist}c@{\dist}c@{\dist}c@{\dist}c@{\dist}c@{\dist}c@{\dist}c@{\dist}c@{\dist}c@{\dist}c@{\dist}c@{\dist}c@{\dist}c@{\dist}c@{\dist}c@{\dist}c@{\dist}c@{\dist}c@{\dist}c@{\dist}c@{\dist}c@{\dist}c@{\dist}c@{\dist}c@{\dist}c@{\dist}}
        	\toprule
        	        & \textbf{Keep}   & \textbf{Keep}  &        &        &        & Super       & Pr    & STM   & Siam       & Siam  &      &      \\
        	        & \textbf{Track}  & \textbf{Track} & CRACT  & TrDiMP & TransT & DiMP        & DiMP  & Track & AttN       & R-CNN & KYS  & DiMP \\
        	        &                 & \textbf{Fast}  & \cite{Fan_2020_arxiv_CRACT} & \cite{Wang_2021_CVPR_TrDiMP} & \cite{Chen_2021_CVPR_TransT} & \cite{Danelljan_2019_github_pytracking} & \cite{Danelljan_2020_CVPR_PRDIMP} & \cite{Fu_2021_CVPR_STMTrack} & \cite{Yu_2020_CVPR_SiamAttN} & \cite{Voigtlaender_2020_CVPR_SiamRCNN} & \cite{Bhat_2020_ECCV_KYS} & \cite{Bhat_2019_ICCV_DIMP}\\          
        	\midrule
        	UAV123  & \best{69.7} & \scnd{69.5} & 66.4        & 67.5        & 69.1 & 68.1 & 68.0 & 64.7        & 65.0 & 64.9 & --   & 65.3 \\
        	OTB-100 & 70.9        & 71.2        & \best{72.6} & 71.1        & 69.4 & 70.1 & 69.6 & \scnd{71.9} & 71.2 & 70.1 & 69.5 & 68.4 \\
        	NFS     & \best{66.4} & 65.3        & 62.5        & \scnd{66.2} & 65.7 & 64.7 & 63.5 & --          & --   & 63.9 & 63.5 & 62.0 \\
            \bottomrule
        \end{tabular}
	}\vspace{1mm}
	\caption{Comparison with state-of-the-art on the OTB-100~\cite{WU_2015_TPAMI_OTB}, NFS~\cite{Galoogahi_2017_ICCV_NFS} and UAV123~\cite{Mueller_2016_ECCV_UAV123} datasets in terms of AUC score. 
	}
	\label{tab:nfs_uav_otb}\vspace{-3mm}%
\end{table}

%% file: sections/discussion_and_conclusion.tex
\section{Conclusion}

We propose a novel tracking pipeline employing a learned target candidate association network in order to track both the target and distractor objects. This approach allows us to propagate the identities of all target candidates throughout the sequence. In addition, we propose a training strategy that combines partial annotations with self-supervision. We do so to compensate for lacking ground-truth correspondences between distractor objects in visual tracking.
We conduct comprehensive experimental validation and analysis of our approach on seven generic object tracking benchmarks and set a new state-of-the-art on six.

%% file: supplementary/supplementary.tex
\input{supplementary/tables/data-mining}

In this supplementary material, we first provide details about training in Sec.~\ref{sup:sec:training} and about inference in Sec.~\ref{sup:sec:inference}. We then report a more detailed analysis of out method in Sec.~\ref{sup:sec:analysis} and provide more detailed results of the experiments shown in the main paper in Sec.~\ref{sup:sec:experiments}.

Furthermore, we want to draw the attention of the reader to the compiled video available on the project page at \mbox{\url{https://github.com/visionml/pytracking}}. The video provides more visual insights about our tracker and compares visually with the baseline tracker SuperDiMP. It shows tracking of distractors and indicates the same object ids in consecutive frames with agreeing color.

\section{Training}
\label{sup:sec:training}

First, we describe the training data generation and sample selection to train the network more effectively. Then, we provide additional details about the training procedure such as training in batches, augmentations and synthetic sample generation. Finally, we briefly summarize the employed network architecture.

\subsection{Data-Mining}
We use the LaSOT~\cite{Fan_2019_CVPR_Lasot} training set to train our target candidate association network. In particular, we split the 1120 training sequences randomly into a \emph{train-train} (1000 sequences) and a \emph{train-val} (120 sequences) set. We run the base tracker on all sequences and store the target classifier score map and the search area on disk for each frame. During training, we use the score map and the search area to extract the target candidates and its features to provide the data to train the target candidate association network.

We observed that many sequences or sub-sequences contain mostly one target candidate with a high target classifier score. Thus, in this cases target candidate association is trivial and learning on these cases will be less effective. Conversely, tracking datasets contain sub-sequences that are very challenging (large motion or appearance changes or many distractors) such that trackers often fail. While these sub-sequences lead to a more effective training they are relatively rare such that we decide to actively search the training dataset.

First, we assign each frame to one of six different categories.
We classify each frame based on four observations about the number of candidates, their target classifier score, if one of the target candidates is selected as target and if this selection is correct, see Tab.~\ref{sup:tab:data-mining}. 
A candidate corresponds to the annotated target object if the spatial distance between the candidate location and center coordinate of the target object is smaller than a threshold.

Assigning each frame to the proposed categories, we observe, that the dominant category is \textsf{D} (70\%) that corresponds to frames with a single target candidate matching the annotated target object.
However, we favour more challenging settings for training. In order to learn distractor associations using self supervision, we require frames with multiple detected target candidates. Category \textsf{H} (18.4\%) corresponds to such frames where in addition the candidate with the highest target classifier score matches the annotated target object. Hence, the base tracker selects the correct candidate as target. Furthermore, category \textsf{G} corresponds to frames where the base tracker was no longer able to track the target because the target classifier score of the corresponding candidate fell bellow a threshold. We favour these frames during training in order to learn continue tracking the target even if the score is low.

Both categories \textsf{J} and \textsf{K} correspond to tracking failures of the base tracker. Whereas in \textsf{K} the correct target is detected but not selected, it is not detected in frames of category \textsf{J}. Thus, we aim to learn from tracking failures in order to train the target candidate association network such that it learns to compensate for tracking failures of the base tracker and corrects them. In particular, frames of category \textsf{K} are important for training because the two candidates with highest target classifier score no longer match such that the network is forced to include other cues for matching. We use frames of category \textsf{J} because frames where the object is visible but not detected contain typically many distractors such that these frames are suitable to learn distractor associations using self-supervised learning.  

To summarize, we select only frames with category \textsf{H}, \textsf{K}, \textsf{J} for self-supervised training and sample them with a ratio of $2:1:1$ instead of $10:2:1$ (ratio in the dataset). We ignore frames from category \textsf{D} during self-supervised training because we require frames with multiple target candidates.
Furthermore, we select sub-sequences of two consecutive frames for partially supervised training. We choose challenging sub-sequences that either contain many distractors in each frame (\textsf{HH}, 350k) or sub-sequences where the base tracker fails and switches to track a distractor (\textsf{HK}, 1001) or where the base tracker is no longer able to identify the target with high confidence (\textsf{HG}, 1380). Again we change the sampling ratio from approximately $350:1:1$ to $10:1:1$ during training. We change the sampling ration in order to use failure cases more often during training than they occur in the training set.

\subsection{Training Data Preparation}

During training we use two different levels of augmentation. First, we augment all features of target candidate to enable self-supervised training with automatically produced ground truth correspondences. In addition, we use augmentation to improve generalization and overfitting of the network. 

When creating artificial features we randomly scale each target classifier score, randomly jitter the candidate location within the search area and apply common image transformations to the original image before extracting the appearance based features for the artificial candidates. In particular, we randomly jitter the brightness, blur the image and jitter the search area before cropping the image to the search area. 

To reduce overfitting and improve the generalization, we randomly scale the target candidate scores for synthetic and real sub-sequences. Furthermore, we remove candidates from the sets $\makeSet{V}'$ and $\makeSet{V}$ randomly in order to simulate newly appearing or disappearing objects.
Furthermore, to enable training in batches we require the same number of target candidates in each frame. Thus, we keep the five candidates with the highest target classifier score or add artificial peaks at random locations with a small score such that five candidates per frame are present. When computing the losses, we ignore these artificial candidates.

\subsection{Architecture Details}

We use the SuperDiMP tracker~\cite{Danelljan_2019_github_pytracking} as our base tracker. SuperDiMP employs the DiMP~\cite{Bhat_2019_ICCV_DIMP} target classifier and the probabilistic bounding-box regression of PrDiMP~\cite{Danelljan_2020_CVPR_PRDIMP}, together with improved training settings. It uses a ResNet-50~\cite{He_2016_CVPR_Resnet} pretrained network as backbone feature extractor. We freeze all parameters of SuperDiMP while training the target candidate association network.
To produce the visual features for each target candidate, we use the third layer ResNet-50 features. In particular, we obtain a $29\times29\times1024$ feature map and feed it into a $2\times2$ convolutional layer which produces the $30\times30\times256$ feature map $\vec{f}$. Note, that the spatial resolution of the target classifier score and feature map agree such that extracting the appearance based features $\vec{f}_i$ for each target candidate $v_i$ at location $\vec{c}_i$ is simplified.

Furthermore, we use a four layer Multi-Layer Perceptron (MLP) to project the target classifier score and location for each candidate in the same dimensional space as $\vec{f}_i$. We use the following MLP structure: $3 \to 32 \to 64 \to 128 \to 256$ with batch normalization. Before feeding the candidate locations into the MLP we normalize it according to the image size.

We follow Sarlin~\etal~\cite{Sarlin_2020_CVPR_SuperGlue} when designing the candidate embedding network. In particular, we use self and cross attention layers in an alternating fashion and employ two layers of each type. In addition, we append a $1\times1$ convolutional layer to the last cross attention layer.
Again, we follow Sarlin~\etal~\cite{Sarlin_2020_CVPR_SuperGlue} for optimal matching and reuse their implementation of the Sinkhorn algorithm and run it for 10 iterations.

\section{Inference}
\label{sup:sec:inference}

In this section we provide the detailed algorithm that describes the object association module (Sec.~\ref{sec:objec_association} in the paper). Furthermore, we explain the idea of search area rescaling at occlusion and how it is implemented. We conclude with additional inference details.

\input{supplementary/algorithm}

\subsection{Object Association Module}

Here, we provide a detailed algorithm describing the object association module presented in the main paper, see Alg.~\ref{sup:alg:object-association}. It contains target candidate to object association and the redetection logic to retrieve the target object after it was lost.

First, we will briefly explain the used notation. Each object can be modeled similar to a \emph{class} in programming. Thus, each object $o$ contains attributes that can be accesses using the $"."$ notation. In particular $(o_j).s$ returns the score attribute $s$ of object $o_j$. In total the object class contains two attribute: the list of scores $s$ and the object-id $id$. Both setting and getting the attribute values is possible.

The algorithm requires the following inputs: the set of target candidates $\makeSet{V}$, the set of detected objects $\makeSet{O}'$ and the object selected as target $\hat{o}$ in the previous frame.
First, we check if a target candidate matches with any previously detected object and verify that the assignment probability is higher than a threshold $\omega=0.75$. If such a match exists, we associate the candidate to the object and append its target classifier score to the scores and add the object to the set of currently visible object $\makeSet{O}$. 
If a target candidate matches none of the previously detected objects, we create a new object and add it to $\makeSet{O}$. Hence, previously detected objects that miss a matching candidate are not included in $\makeSet{O}$.
Once, all target candidates are associated to an already existing or newly created object. We check if the object previously selected as target is still visible in the current scene and forms the new target $\hat{o}$. After the object was lost it is possible that the object selected as target is in fact a distractor. Thus, we select an other object as target if this other object achieves a higher target classifier score in the current frame than any score the currently selected object achieved in the past. 
Furthermore, if the object previously selected as target object is no longer visible, we try to redetect it by checking if the object with highest target classifier score in the current frame achieves a score higher than a threshold $\eta=0.25$. If the score is high enough, we select this object as the target.

\subsection{Search Area Rescaling at Occlusion}

The target object often gets occluded or moves out-of-view in many tracking sequences. Shortly before the target is lost the tracker typically detects only a small part of the target object and estimates a smaller bounding box than in the frames before. The used base tracker SuperDiMP employs a search area that depends on the currently estimated bounding box size. Thus, a partially visible target object causes a small bounding box and search area. The problem of a small search area is that it complicates redetecting the target object, \eg, the target reappears at a slightly different location than it disappeared and if the object then reappears outside of the search area redetection is prohibited.
Smaller search areas occur more frequently when using the target candidate association network because it allows to track the object longer until we declare it as lost.

Hence, we use a procedure to increase the search area if it decreased before the target object was lost.
First, we store all search are resolutions during tracking in an list $a$ as long as the object is detected. If the object was lost $k$ frames ago, we compute the new search area by averaging the last $k$ entries of $a$ larger than the search area at occlusion. We average at most over 30 previous search areas to compute the new one. If the target object was not redetected within these 30 frames with keep the search area fixed until redetection.  

\input{supplementary/tables/abblation}
\input{supplementary/tables/tca_ablation}

\input{supplementary/figures/lasot_plots}
\input{supplementary/tables/lasot}

\subsection{Inference Details}

In contrast to training, we use all extracted target candidates to compute the candidate associations between consecutive frames. In order to save computations, we extract the candidates and features only for the current frame and cache the results such that they can be reused when computing the associations in the next frames.

\subsubsection{KeepTrack Settings}

We use the same settings as for SuperDiMP but increase the search area scale from 6 to 8 leading to a larger search are (from $352\times352$ to $480\times480$) and to a larger target score map (from $22\times22$ to $30\times30$). In addition, we employ the aforementioned search area rescaling at occlusion and skip running the target candidate association network if only one target candidates with high target classifier score is detected in the current and previous frame, in order to save computations. 

\subsubsection{KeepTrackFast Settings}

We use the same settings as for SuperDiMP. In particular, we keep the search area scale and target score map resolution the same to achieve a faster run-time. In addition, we reduce the number of bounding box refinement steps from 10 to 3 which reduces the bounding box regression time significantly. Moreover, we double the target candidate extraction threshold $\tau$ to 0.1. This step ensures that we neglegt local maxima with low target classifier scores and thus leads to less frames with multiple detected candidates. Hence, \emph{KeepTrackFast} runs the target candidate association network less often than \emph{KeepTrack}.

\input{supplementary/figures/lasot_ext_sub_plots}

\section{More Detailed Analysis}
\label{sup:sec:analysis}

In addition to the ablation study presented in the main paper (Sec.~\ref{sec:ablation_study}) we provide more settings in order to assess the contribution of each component better. In particular, we split the term \emph{search area adaptation} into \emph{larger search area} and \emph{search area rescaling}. Where larger search area refers to a search area scale of 8 instead of 6 and a search area resolution of 480 instead of 352 in the image domain. Tab.~\ref{sup:tab:component_ablation} shows all the results on NFS~\cite{Galoogahi_2017_ICCV_NFS}, UAV123~\cite{Mueller_2016_ECCV_UAV123} and LaSOT~\cite{Fan_2019_CVPR_Lasot}. We run each experiment five times and report the average. We conclude that both search area adaptation techniques improve the tracking quality but we achieve the best results on all three datasets when employing both at the same time.
Furthermore, we evaluate the target candidate association network with different numbers of Sinkhorn iterations and with different number of GNN layers of the embedding network or dropping it at all, see Tab.~\ref{sup:tab:tac_ablation}. We conclude, that using the target candidate association network even without any GNN layers outperforms the baseline on all three datasets. 
In addition, using either two or nine GNN layers improves the performance even further on all datasets. We achieve the best results when using nine GNN layers and 50 Sinkhorn iterations. However, using a large candidate embedding network and a high number of Sinkhorn iterations reduces the run-time of the tracker to 12.7 FPS. Hence, using only two GNN layers and 10 Sinkhorn iterations results in a negligible decrease of 0.1 on UAV123 and LaSOT but accelerates the run-time by $44\%$. 

\input{supplementary/figures/failure_cases}

\section{Experiments}\label{sup:sec:experiments}

We provide more details to complement the state-of-the-art comparison performed in the paper. And provide results for the VOT2018LT~\cite{Kristan_2020_ECCVW_VOT2020} challenge.

\subsection{LaSOT and LaSOTExtSub}
In addition to the success plot, we provide the normalized precision plot on the LaSOT~\cite{Fan_2019_CVPR_Lasot} test set (280 videos) and LaSOTExtSub~\cite{Fan_2019_CVPR_Lasot} test set (150 videos). The normalized precision score $\text{NPr}_D$ is computed as the percentage of frames where the normalized distance (relative to the target size) between the predicted and ground-truth target center location is less than a threshold $D$. $\text{NPr}_D$ is plotted over a range of thresholds $D \in [0,0.5]$. The trackers are ranked using the AUC, which is shown in the legend.
Figs.~\ref{sup:fig:lasot_norm_prec} and~\ref{sup:fig:lasot_ext_sub_norm_prec} show the normalized precision plots.
We compare with state-of-the-art trackers and report their success (AUC) in Tab.~\ref{sup:tab:lasot} and where available we show the raw results in Fig.~\ref{sup:fig:lasot}. In particular, we use the raw results provided by the authors except for DaSiamRPN~\cite{Zhu_2018_ECCV_DaSiamRPN}, GlobaTrack~\cite{Huang_2020_AAAI_GlobalTrack}, SiamRPN++~\cite{Li_2019_CVPR_SiamRPN++} and SiamMask~\cite{Wang_2019_CVPR_SiamMask} such results were not provided such that we use the raw results produced by Fan~\etal~\cite{Fan_2019_CVPR_Lasot}. Thus, the exact results for these methods might be different in the plot and the table, because we show in the table the reported result the corresponding paper. Similarly, we obtain all results on LaSOTExtSub directly from Fan~\etal~\cite{Fan_2020_IJCV_Lasot_ext} except the result of SuperDiMP that we produced.

\input{supplementary/figures/otb_nfs_uav_success_plots}
\input{supplementary/tables/otb_nfs_uav}

\input{supplementary/tables/vot2018lt}

\input{supplementary/tables/uav_attributes}
\input{supplementary/tables/lasot_attributes}

\subsection{UAV123, OTB-100 and NFS}
We provide the success plot over the 123 videos of the UAV123 dataset~\cite{Mueller_2016_ECCV_UAV123} in Fig.~\ref{sup:fig:uav}, the 100 videos of the OTB-100 dataset~\cite{WU_2015_TPAMI_OTB} in Fig.~\ref{sup:fig:otb} and the 100 videos of the NFS dataset~\cite{Galoogahi_2017_ICCV_NFS} in Fig.~\ref{sup:fig:nfs}. We compare with state-of-the-art trackers SuperDiMP~\cite{Danelljan_2019_github_pytracking}, PrDiMP50~\cite{Danelljan_2020_CVPR_PRDIMP}, UPDT~\cite{Bhat_2018_ECCV_UPDT}, SiamRPN++~\cite{Li_2019_CVPR_SiamRPN++}, ECO~\cite{Danelljan_2017_CVPR_ECO}, DiMP~\cite{Bhat_2019_ICCV_DIMP}, CCOT~\cite{Danelljan_2016_ECCV_CCTO}, MDNet~\cite{Nam_2016_CVPR_MDNet}, ATOM~\cite{Danelljan_2019_CVPR_ATOM}, and DaSiamRPN~\cite{Zhu_2018_ECCV_DaSiamRPN}. Our method provides a significant gain over the baseline SuperDiMP on UAV123 and NFS and performs among the top methods on OTB-100. Tab.~\ref{sup:tab:otb_nfs_uav} shows additional results on UAV123, OTB-100 and NFS in terms of success (AUC).

\subsection{VOT2018LT~\cite{Matej_2018_ECCVW_VOT2018}}
Next, we evaluate our tracker on the 2018 edition of the VOT~\cite{Kristan_2016_TPAMI_VOT} long-term tracking challenge. We compare with the top methods in the challenge~\cite{Matej_2018_ECCVW_VOT2018}, as well as more recent methods. The dataset contains 35 videos with 4200 frames per sequence on average. Trackers are required to predict a confidence score that the target is present in addition to the bounding box for each frame. Trackers are ranked by the F-score, evaluated for a range of confidence thresholds. As shown in Tab.~\ref{sup:tab:vot2018lt}, our tracker achieves the best results in all three metrics and outperforms the base tracker SuperDiMP by almost $10\%$ in F-score.

\section{Speed Analysis}

Our method adds an overhead of 19.3ms compared to the baseline tracker. Whereas target candidate extraction is required for every frame, running the target candidate association network is only required if more than one candidate is detected. Candidate extraction is relatively fast and takes only 2.7~ms while performing candidate association requires 16.6~ms. Where computing the candidate embedding takes 10.0~ms, running the Sinkhorn algorithm for 10 iterations takes 3.5~ms and object association 3.1~ms. Thus, we achieve an average run-time of 18.3~FPS for \emph{KeepTrack} and 29.6~FPS for \emph{KeepTrackFast} when using SuperDiMP as base tracker. We report this timings on a singe Nvidia GTX 2080Ti GPU. 

\subsection{Number of Candidates and Time Complexity}
In practice, we found the number of target candidates to vary from 0 up to 15 in exceptional cases. For frames with less than 2 candidates, we naturally do not need to apply our association module. Further, we observed no measurable increase in run-time from 2 to 15 candidates, due to the effective parallelization. Therefore, we do not explicitly limit the number of detected candidates.

\section{Failure Cases}

While \emph{KeepTrack} is particularly powerful when distractor objects appear in the scene, it also fails to track the target object in complex scenes, such as the examples shown in Fig.~\ref{sup:fig:failure_cases}.

The top row shows such a challenging case, where the target object is the right hand of the person on the right (indicated by [\bbox{_green}] or [\bbox{_red}]). \emph{KeepTrack} manages to individually track all hands (\object{_coral},\object{_green},\object{_yellow}) visible in the search area until frame number 134. In the next frame, both hands of the person are close and our tracker only detects one candidate for both hands (\object{_green}). Thus, the tracker assigns the target id to the remaining candidate. The tracker detects two candidates (\object{_green},\object{_skyblue}) as soon as both hands move apart in frame 143. However, now it is unclear which hand is the sought target. If two objects approach each other it is unclear whether they cross each other or not. In this scenario positional information is of limited use. Hence, deeper understanding of the scene and the target object seems necessary to mitigate such failure cases.

The bottom row in Fig.~\ref{sup:fig:failure_cases} shows a similar failure case where again a distractor object (\object{_skyblue}) crosses the target's (\object{_green}) location. This time, the tracker fails to extract the candidate corresponding to the target from frame number 77 on wards. The tracker detects that the target candidate previously assumed to represent the target has vanished but the remaining distractor object (\object{_skyblue}) achieves such a high target score that the tracker reconsiders its previous target selection and continues tracking the distractor object (\object{_skyblue}) instead. Thus, the tracker continues tracking the distractor object even if the a target candidate for the sought target (\object{_violet}) appears (frame number 93).

\section{Attributes}

Tabs.~\ref{sup:tab:uav_attributes} and~\ref{sup:tab:lasot_attributes} show the results of various trackers including \emph{KeepTrack} and \emph{KeepTrackFast} based on different sequence attributes. We observe that both trackers are superior to other trackers on UAV123 for most attributes. In particular, we outperform the runner-up by a large margin in terms of AUC on the sequences corresponding to the following attributes: 
\emph{Aspect Ratio Change} ($+2.5/2.6\%$), Full Occlusion ($+1.8/1.9\%$), Partial Occlusion ($+2.5/2.3\%$), Background Clutter ($+1.5/1.3\%$), Illumination Variation ($+1.7/1.5\%$), Similar Object ($+1.8/1.1\%$).
Especially, the superior performance on sequences with the attributes \emph{Full Occlusion}, \emph{Partial Occlusion}, \emph{Background Clutter} and \emph{Similar Object} clearly demonstrates that \emph{KeepTrack} mitigates the harmful effect of distractors and allows to track the target object longer and more frequently than other trackers. Fig.~\ref{sup:fig:uav} shows a similar picture: \emph{KeepTrack} is the most robust tracker but others achieve a higher bounding box regression accuracy.
In addition, Tab.~\ref{sup:tab:uav_attributes} reveals that \emph{KeepTrack} achieves the highest (\textcolor{red}{red}) or second-highest (\textcolor{blue}{blue}) AUC on the sequences corresponding to each attribute except for the attribute \emph{Out-of-View}.

The attribute-based analysis on LaSOT allows similar observations. In particular, \emph{KeepTrack} and \emph{KeepTrackFast} outperform all other trackers by a large margin in AUC on the sequences corresponding oth the following attributes: \emph{Partial Occlusion} ($+1.8/1.5\%$), \emph{Background Clutter} ($+2.3/1.2$, \emph{Viewpoint Change} ($+1.6/2.3\%$), \emph{Full Occlusion} ($+2.7/1.8\%$), \emph{Fast Motion} ($4.1/3.5\%$), \emph{Out-of-View} ($+1.9/1.2\%$), \emph{Low Resolution} ($+3.4/3.4\%$). Moreover, the superior performance is even clearer when comparing to the base tracker SuperDiMP, \eg, and improvement of $+6.0/5.2\%$ for \emph{Full Occlusion} or $+7.0/6.4\%$ for \emph{Fast Motion}. \emph{KeepTrack} achieves the highest AUC score for every attributed except two where \emph{KeepTrackFast} achieves slightly higher scores.
Again, the best performance on sequences with attributes such as \emph{Background Clutter} or \emph{Full Occlusion} clearly demonstrates the effectiveness of our proposed target and distractor association strategy. 

%% file: supplementary/tables/data-mining.tex
\begin{table}[b]
	\centering
	\newcommand{\best}[1]{\textbf{#1}}
	\newcommand{\dist}{\hspace{5pt}}%
	\newcommand{\yes}{\textcolor{black}{\checkmark}}
    \newcommand{\no}{\textcolor{black}{x}}
	\resizebox{1.0\columnwidth}{!}{%
        \begin{tabular}{c@{\dist}c@{\dist}c@{\dist}c@{\dist}c@{\dist}c@{\dist}c@{\dist}}
        	\toprule
        	         
        	       &            & Is a candidate        & Does the candidate & Does any      &        & \\
        	       & Number     & selected              & with max score     & candidate     &        & \\
        	       & of         & as target?            & correspond         & correspond to & Num    & \\
        	Name   & candidates & $\max(s_i) \geq \eta$ & to the target?     & the target?   & Frames & Ratio\\

        	\midrule
        	 \textsf{D}    & $1$    & \yes &  \yes & \yes & 1.8M & 67.9\% \\
        	 \textsf{H}    & $>1$   & \yes &  \yes & \yes & 498k & 18.4\% \\
        	 \textsf{G}    & $>1$   & \no  &  \yes & --   &   8k &  0.3\% \\
        	 \textsf{J}    & $>1$   & \yes &  \no  & \no  &  76k &  2.8\% \\
        	 \textsf{K}    & $>1$   & \yes &  \no  & \yes &  42k &  1.5\% \\
        	 \textsf{other}& --     & --   &  --   & --   & 243k &  9.1\% \\\bottomrule
        \end{tabular}
	}
	\caption{Categories and specifications for each frame in the training dataset used for data-mining.}
	\label{sup:tab:data-mining}%
\end{table}

%% file: supplementary/algorithm.tex
\begin{algorithm}[t]
    \newcommand{\assign}{\leftarrow}
    \newcommand{\algcomment}[1]{\hfill\textit{#1}}
    \newcommand{\target}{\hat{o}}
    \newcommand{\id}{\mathrm{id}}
    \newcommand{\dustbin}{\mathrm{dustbin}}
	\caption{Object Association Algorithm.}
	\begin{algorithmic}[1]
		\Require Set of target candidates $\makeSet{V}$
		\Require Set of objects of previous frame $\makeSet{O}'$
		\Require Target object $\target'$
        \State $\makeSet{O} = \{\}$, $N = |\makeSet{V}|$    
		\For{$i = 1, \ldots, N$} 
			\If{$\mathrm{matchOf}(v_i) \neq \dustbin$ \&\& $p(v_i) \geq \omega$}
			    \State $v'_{j} \assign \mathrm{matchOf}(v_i)$
			    \State $(o'_j).s \assign \mathrm{concat}((o'_j).s, [s_i])$
                \State $o_i \assign o'_j$
            \Else
                \State $o_i \assign \mathrm{new\, Object}(\mathrm{getNewId}(),[s_i])$   
			\EndIf
			
		    \State $\makeSet{O} \assign \makeSet{O} \cup \{o_i\}$ 
		\EndFor
		
		\If{$\target' \neq \mathrm{none}$ \textbf{and} $\target'.\id \in \{o.\id \mid o\in \makeSet{O}\}$}
		    \State $\target = \mathrm{getObjectById}(\makeSet{O}, \target'.\id)$
			\For{$i = 1, \ldots, N$}
			    \If{$\max(\target.s) < (o_i).s[-1]$}
			        \State $\target = o_i$ 
			    \EndIf
	        \EndFor
        \Else
            \State $i = \mathrm{argmax}_i\{(o_i).s[-1]) \mid o_i \in \makeSet{O}\}$ 
            \If{$(o_i).s[-1] \geq \eta$}
               \State $\target = o_i$
            \Else
                \State $\target = \mathrm{none}$
            \EndIf
        \EndIf
        \State\Return{$\target$, $\makeSet{O}$}
	\end{algorithmic}
	\label{sup:alg:object-association}
\end{algorithm}

%% file: supplementary/tables/abblation.tex
\begin{table}[t]
	\centering
	\newcommand{\best}[1]{\textbf{#1}}
	\newcommand{\dist}{\hspace{5pt}}%
	\newcommand{\yes}{\textcolor{black}{\checkmark}}
	\resizebox{1\columnwidth}{!}{%
        \begin{tabular}{c@{\dist}c@{\dist}c@{\dist}c@{\dist}|c@{\dist}c@{\dist}c@{\dist}}
        	\toprule
        	         
        	         Memory     & Larger  & Search    & Candidate   &  \\
        	         Memory     & Search  & Area      & Association &  \\
        	         Confidence & Area    & Rescaling & Network     & NFS & UAV123 & LaSOT  \\
        	\midrule
        	 --   & --   & --   & --   & 64.4 & 68.2 & 63.5 \\
        	 \yes & --   & --   & --   & 64.7 & 68.0 & 65.0 \\
         	 \yes & \yes & --   & --   & 65.3 & 68.4 & 65.5 \\
         	 \yes & --   & \yes & --   & 64.7 & 68.4 & 65.8 \\
        	 \yes & \yes & \yes & --   & 65.2 & 69.1 & 65.8 \\
        	 \yes & \yes & \yes & \yes & \best{66.4} & \best{69.7} & \best{67.1} \\\bottomrule
        \end{tabular}
	}\vspace{1mm}%
	\caption{Impact of each component in terms of AUC (\%) on three datasets. The first row corresponds to our SuperDiMP baseline.
	}
	\label{sup:tab:component_ablation}%
\end{table}

%% file: supplementary/tables/tca_ablation.tex
\begin{table}[t]
	\centering
	\newcommand{\best}[1]{\textbf{#1}}
	\newcommand{\dist}{\hspace{10pt}}%
	\newcommand{\yes}{\textcolor{black}{\checkmark}}
	\resizebox{0.9\columnwidth}{!}{%
        \begin{tabular}{c@{\dist}c@{\dist}|c@{\dist}c@{\dist}c@{\dist}|c@{\dist}}
        	\toprule
        	 Num GNN  & Num Sinkhorn &     &        &       &     \\
        	 Layers   & iterations   & NFS & UAV123 & LaSOT & FPS \\
        	\midrule
        	-- & -- & 65.2 & 69.1 & 65.8 & -- \\
        	0  & 50 & 65.9 & 69.2 & 66.6 & -- \\
        	2  & 10 & 66.4 & 69.7 & 67.1 & 18.3\\
        	9  & 50 & 66.4 & 69.8 & 67.2 & 12.7\\\bottomrule
        \end{tabular}
	}\vspace{1mm}%
	\caption{Impact of each component of the Target Candidate Association Network in terms of AUC (\%) on three datasets.
	}\vspace{0mm}
	\label{sup:tab:tac_ablation}%
\end{table}

%% file: supplementary/figures/lasot_plots.tex
\begin{figure*}[t]
	\centering
	\newcommand{\wid}{0.45\linewidth}%
	\subfloat[Success\label{sup:fig:lasot_success}]{\includegraphics[width=\wid]{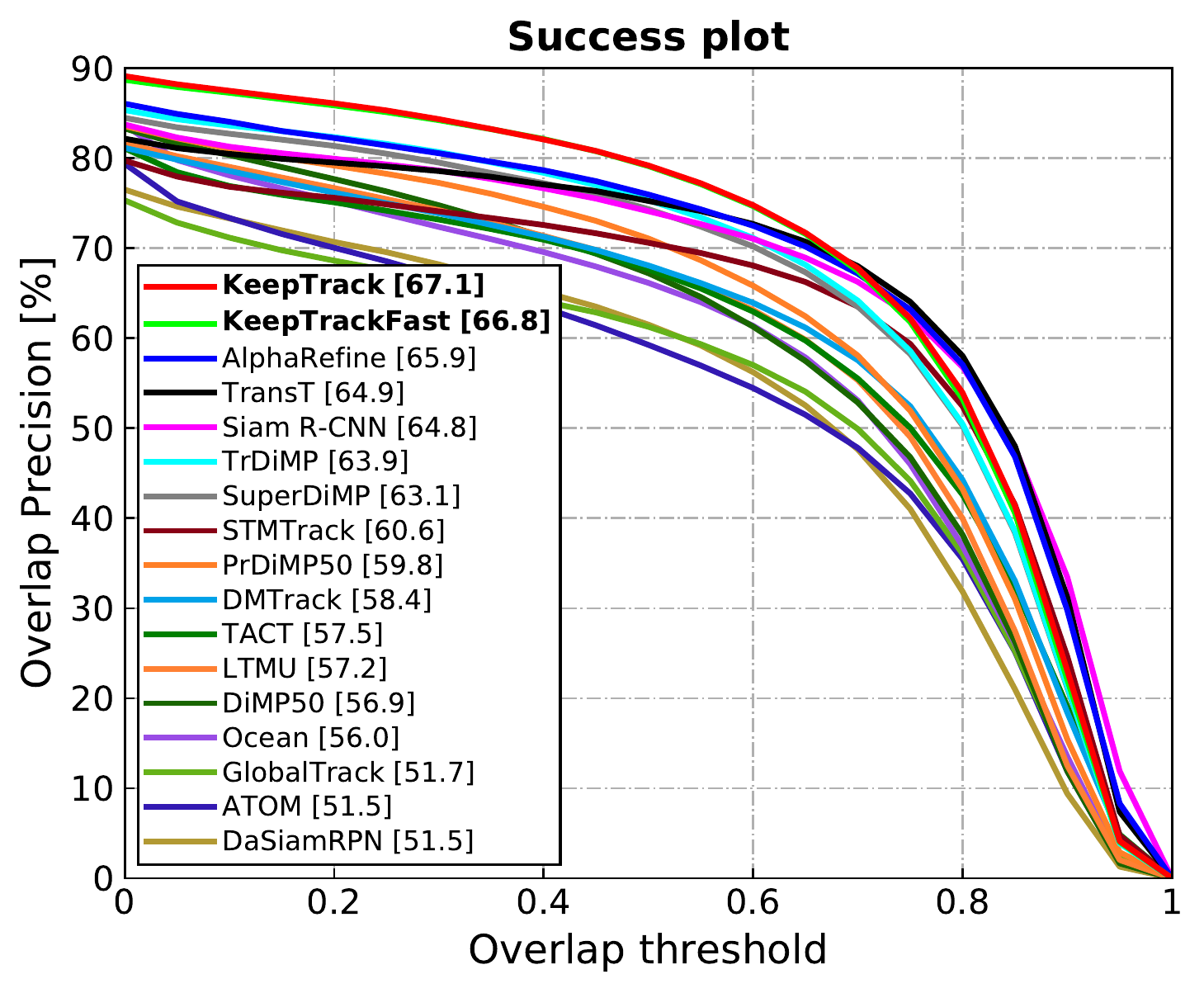}}
	\subfloat[Normalized Precision\label{sup:fig:lasot_norm_prec}]{\includegraphics[width = \wid]{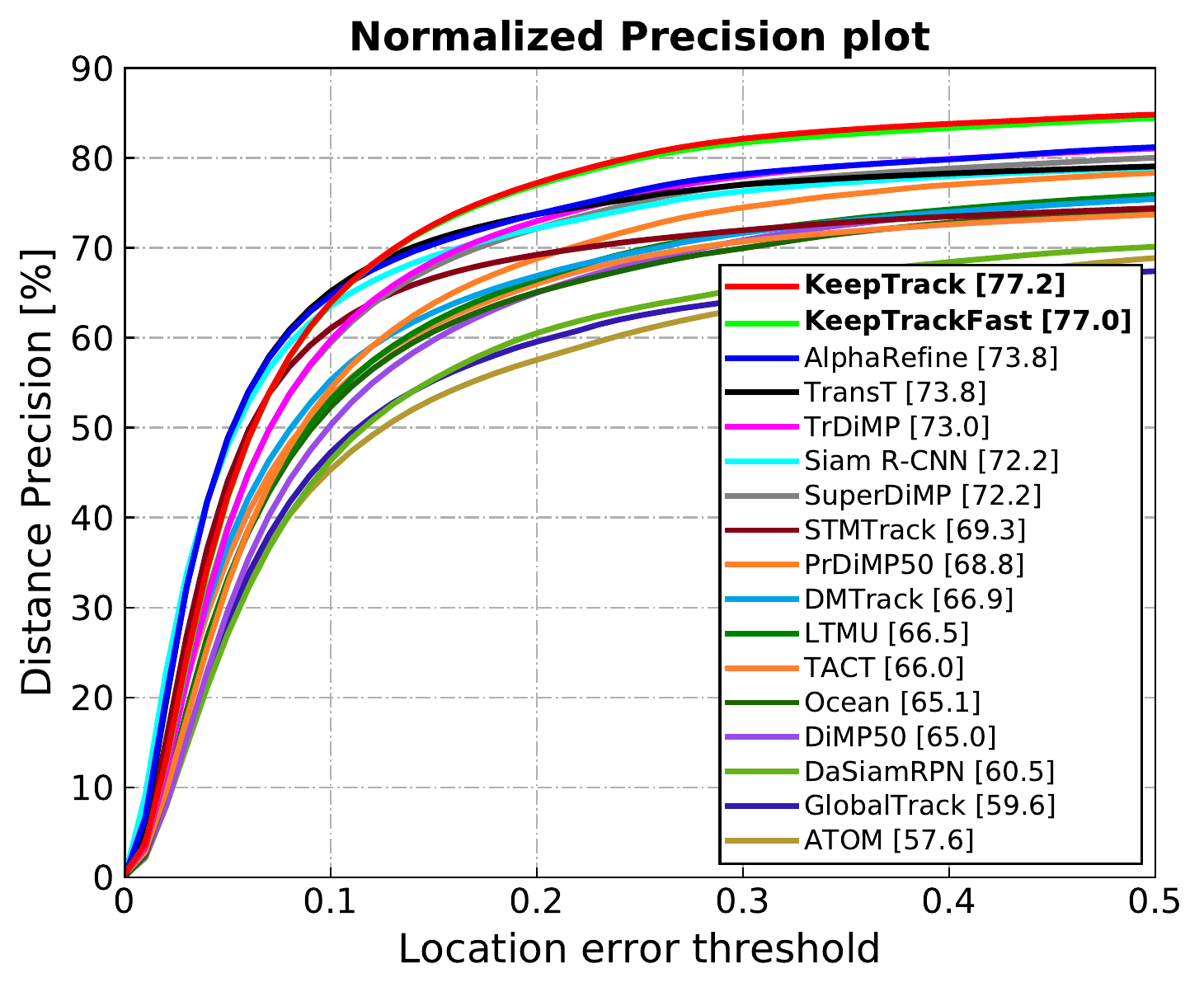}}%
	\caption{Success and normalized precision plots on LaSOT~\cite{Fan_2019_CVPR_Lasot}. Our approach outperforms all other methods by a large margin in AUC, reported in the legend.}%
	\label{sup:fig:lasot}
\end{figure*}

%% file: supplementary/tables/lasot.tex
\begin{table*}[t]
	\centering
	\newcommand{\best}[1]{\textbf{\textcolor{red}{#1}}}
	\newcommand{\scnd}[1]{\textbf{\textcolor{blue}{#1}}}
	\newcommand{\dist}{\hspace{5pt}}%
	\resizebox{1.00\textwidth}{!}{%
        \begin{tabular}{l@{\dist}c@{\dist}c@{\dist}c@{\dist}c@{\dist}c@{\dist}c@{\dist}c@{\dist}c@{\dist}c@{\dist}c@{\dist}c@{\dist}c@{\dist}c@{\dist}c@{\dist}c@{\dist}c@{\dist}c@{\dist}c@{\dist}c@{\dist}c@{\dist}c@{\dist}c@{\dist}c@{\dist}c@{\dist}c@{\dist}c@{\dist}c@{\dist}c@{\dist}c@{\dist}c@{\dist}c@{\dist}c@{\dist}c@{\dist}c@{\dist}c@{\dist}}
        	\toprule
        	        & \textbf{Keep}  & \textbf{Keep}  & Alpha  &        & Siam  & Tr   & Super & STM   & Pr   & DM    &      &      &      &      &       & Siam \\ 
        	        & \textbf{Track} & \textbf{Track} & Refine & TransT & R-CNN & DiMP & Dimp  & Track & DiMP & Track & TLPG & TACT & LTMU & DiMP & Ocean & AttN \\
        	        & & \textbf{Fast} & \cite{Yan_2021_CVPR_AlphaRefine} & \cite{Chen_2021_CVPR_TransT} & \cite{Voigtlaender_2020_CVPR_SiamRCNN} & \cite{Wang_2021_CVPR_TrDiMP} & \cite{Danelljan_2019_github_pytracking} & \cite{Fu_2021_CVPR_STMTrack} & \cite{Danelljan_2020_CVPR_PRDIMP} & \cite{Zhang_2021_CVPR_DMTrack} & \cite{Li_2020_IJCAI_TLPG} & \cite{Choi_2020_ACCV_TACT} & \cite{Dai_2020_CVPR_LTMU} & \cite{Bhat_2019_ICCV_DIMP} & \cite{Zhang_2020_ECCV_Ocean} & \cite{Yu_2020_CVPR_SiamAttN} \\
        	\midrule
        	LaSOT  & \best{67.1} & \scnd{66.8} & 65.3 & 64.9 & 64.8 & 63.9 & 63.1 & 60.6 & 59.8 & 58.4 & 58.1 & 57.5 & 57.2 & 56.9 & 56.0  & 56.0 \\ \bottomrule
        	\toprule
        	&       & Siam & Siam  & PG  & FCOS & Global &      & DaSiam & Siam & Siam &       & Siam   & Retina & Siam &        &     \\
        	& CRACT & FC++ & GAT   & NET & MAML & Track  & ATOM & RPN    & BAN  & CAR  & CLNet & RPN++  & MAML   & Mask & ROAM++ & SPLT\\
        	& \cite{Fan_2020_arxiv_CRACT} & \cite{Xu_2020_AAAI_SiamFCpp} & \cite{Guo_2020_arxiv_SiamGAT} & \cite{Bingyan_2020_ECCV_PGNet} & \cite{Wang_2020_CVPR_MAML} & \cite{Huang_2020_AAAI_GlobalTrack} & \cite{Danelljan_2019_CVPR_ATOM} & \cite{Zhu_2018_ECCV_DaSiamRPN}$^\dagger$ & \cite{Chen_2020_CVPR_SiamBAN} & \cite{Guo_2020_CVPR_SiamCAR} & \cite{Dong_2020_ECCV_CLNet} & \cite{Li_2019_CVPR_SiamRPN++}$^\dagger$ & \cite{Wang_2020_CVPR_MAML} & \cite{Wang_2019_CVPR_SiamMask}$^\dagger$ & \cite{Yang_2020_CVPR_ROAM} & \cite{Yan_2019_ICCV_SPLT}\\
        	\midrule
        	LaSOT  & 54.9  & 54.4 & 53.9  & 53.1   & 52.3 & 52.1   & 51.5 & 51.5 & 51.4 & 50.7 & 49.9  & 49.6   & 48.0   & 46.7 & 44.7 & 42.6\\\bottomrule
        \end{tabular}
	}
	\caption{Comparison with state-of-the-art on the LaSOT~\cite{Fan_2019_CVPR_Lasot} test set in terms of overall AUC score. The average value over 5 runs is reported for our approach. The symbol $^\dagger$ marks results that were produced by Fan~\etal~\cite{Fan_2019_CVPR_Lasot} otherwise they are obtained directly from the official paper.}
	\label{sup:tab:lasot}%
\end{table*}

%% file: supplementary/figures/lasot_ext_sub_plots.tex
\begin{figure*}[!t]
	\centering
	\newcommand{\wid}{0.44\textwidth}%
	\subfloat[Success\label{sup:fig:lasot_ext_sub_success}]{\includegraphics[width=\wid]{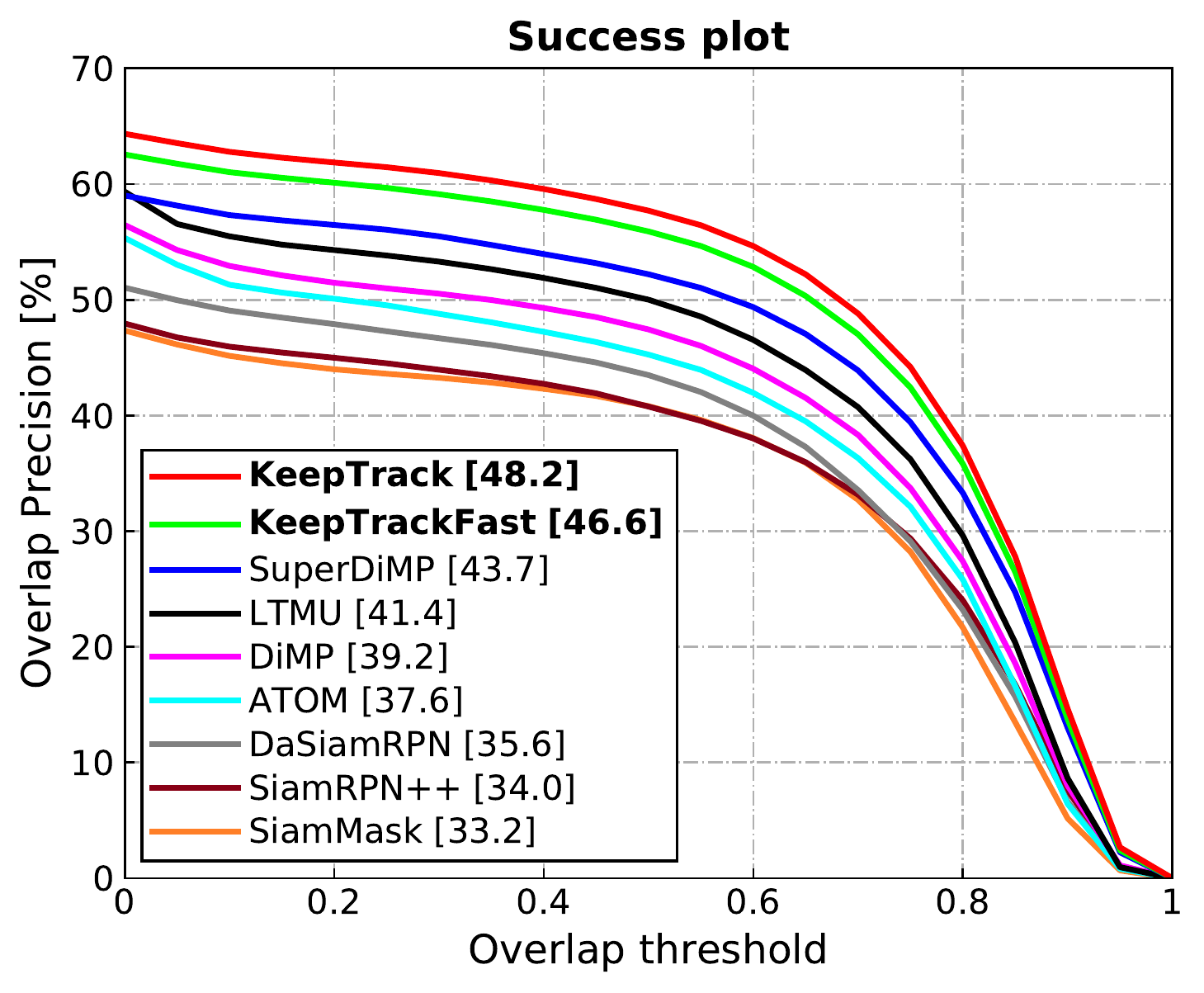}}~%
	\subfloat[Normalized Precision\label{sup:fig:lasot_ext_sub_norm_prec}]{\includegraphics[width = \wid]{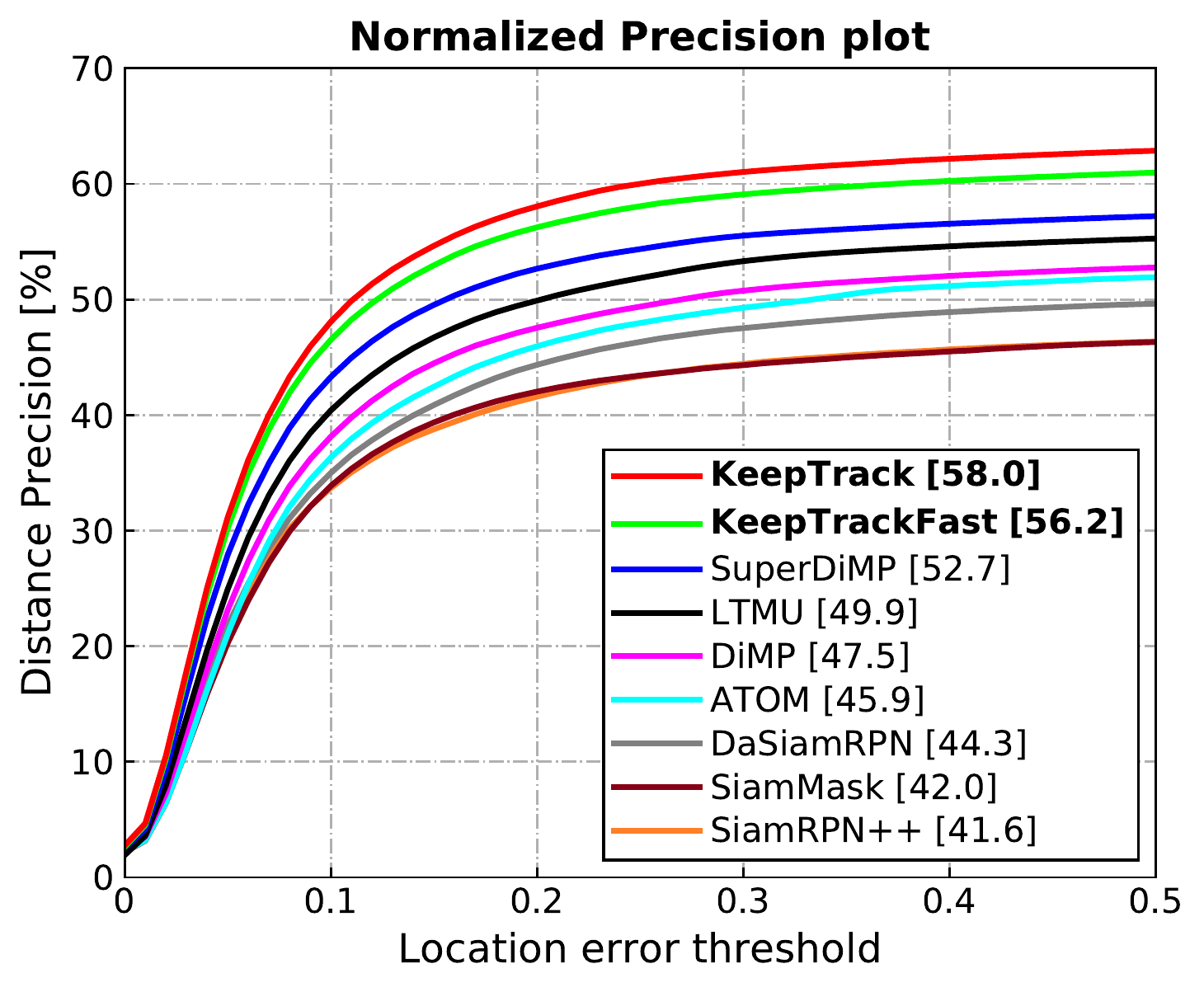}}%
	\caption{Success and normalized precision plots on LaSOTExtSub~\cite{Fan_2020_IJCV_Lasot_ext}. Our approach outperforms all other methods by a large margin.}%
	\label{sup:fig:lasot_ext_sub}%
\end{figure*}

%% file: supplementary/figures/failure_cases.tex
\begin{figure*}[t]
\centering
\includegraphics[width=\textwidth, keepaspectratio]{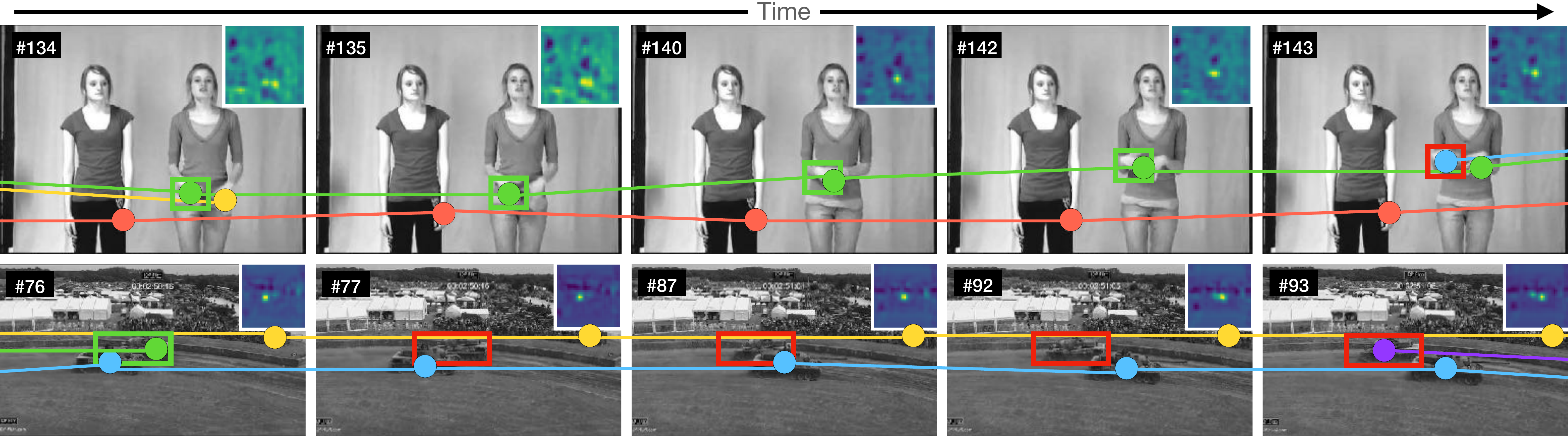}
\caption{Failure Cases: a very challenging case is when a distractor crosses the target’s location, since positional information is then of limited use. The box represents the ground truth bounding box of the target object, where green indicates the the selected target candidates corresponds to the sought target and red indicates that the tracker selected a candidate corresponding to a distractor object.}\label{sup:fig:failure_cases}
\end{figure*}

%% file: supplementary/figures/otb_nfs_uav_success_plots.tex
\begin{figure*}[!t]
	\centering%
	\newcommand{\wid}{0.33\textwidth}%
	\subfloat[UAV123\label{sup:fig:uav}]{\includegraphics[width=\wid]{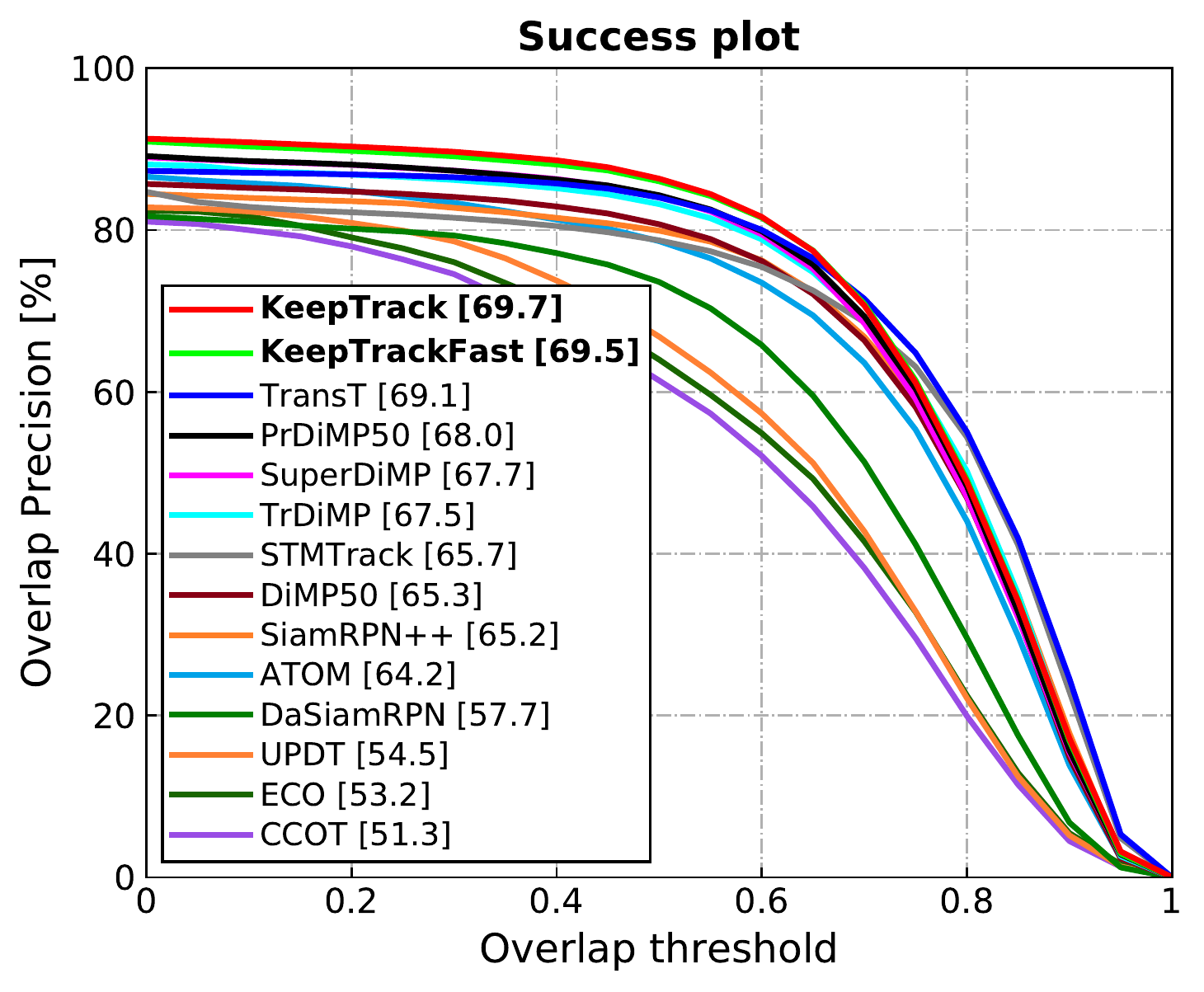}}~%
	\subfloat[OTB-100\label{sup:fig:otb}]{\includegraphics[width = \wid]{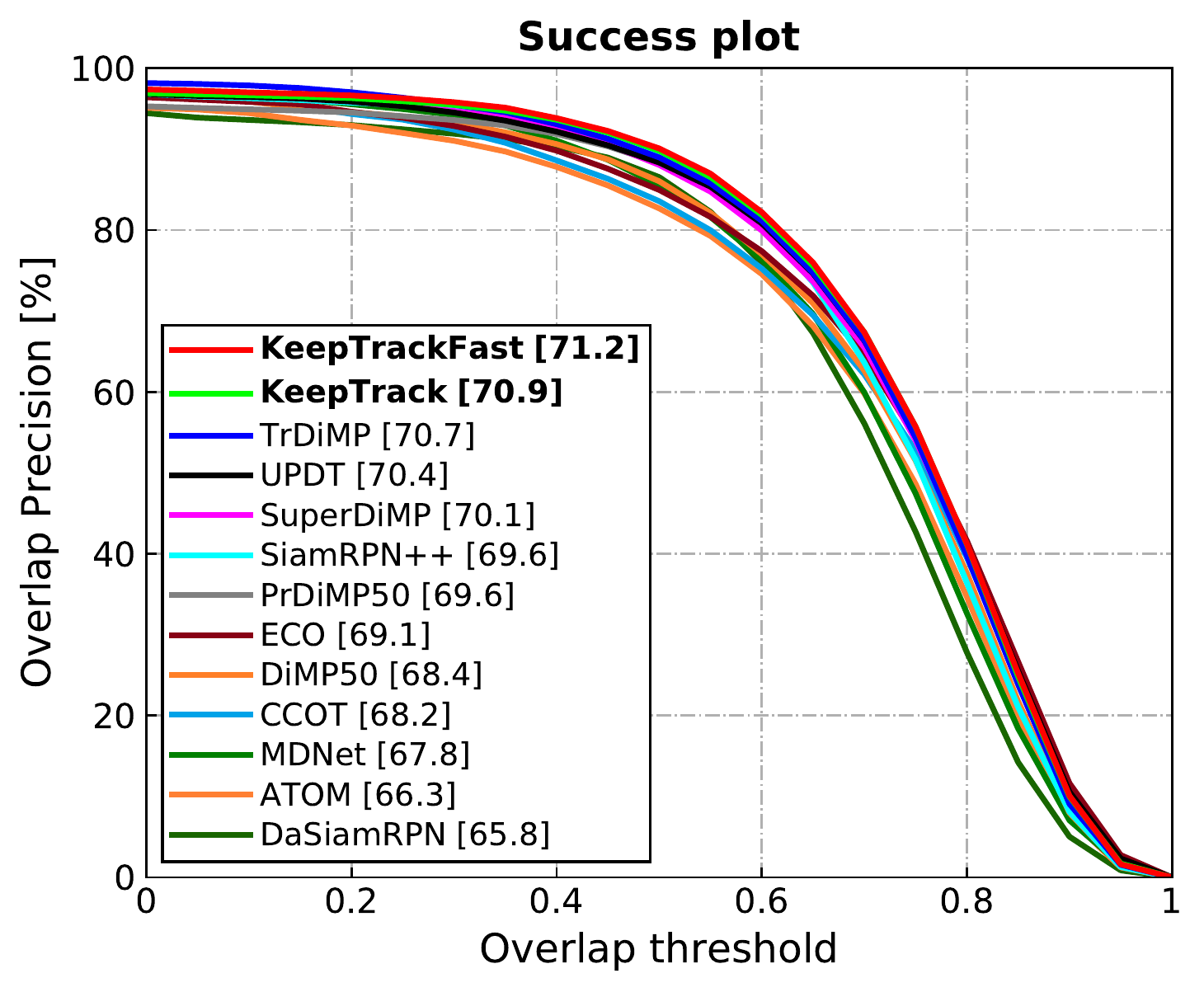}}%
	\subfloat[NFS\label{sup:fig:nfs}]{\includegraphics[width = \wid]{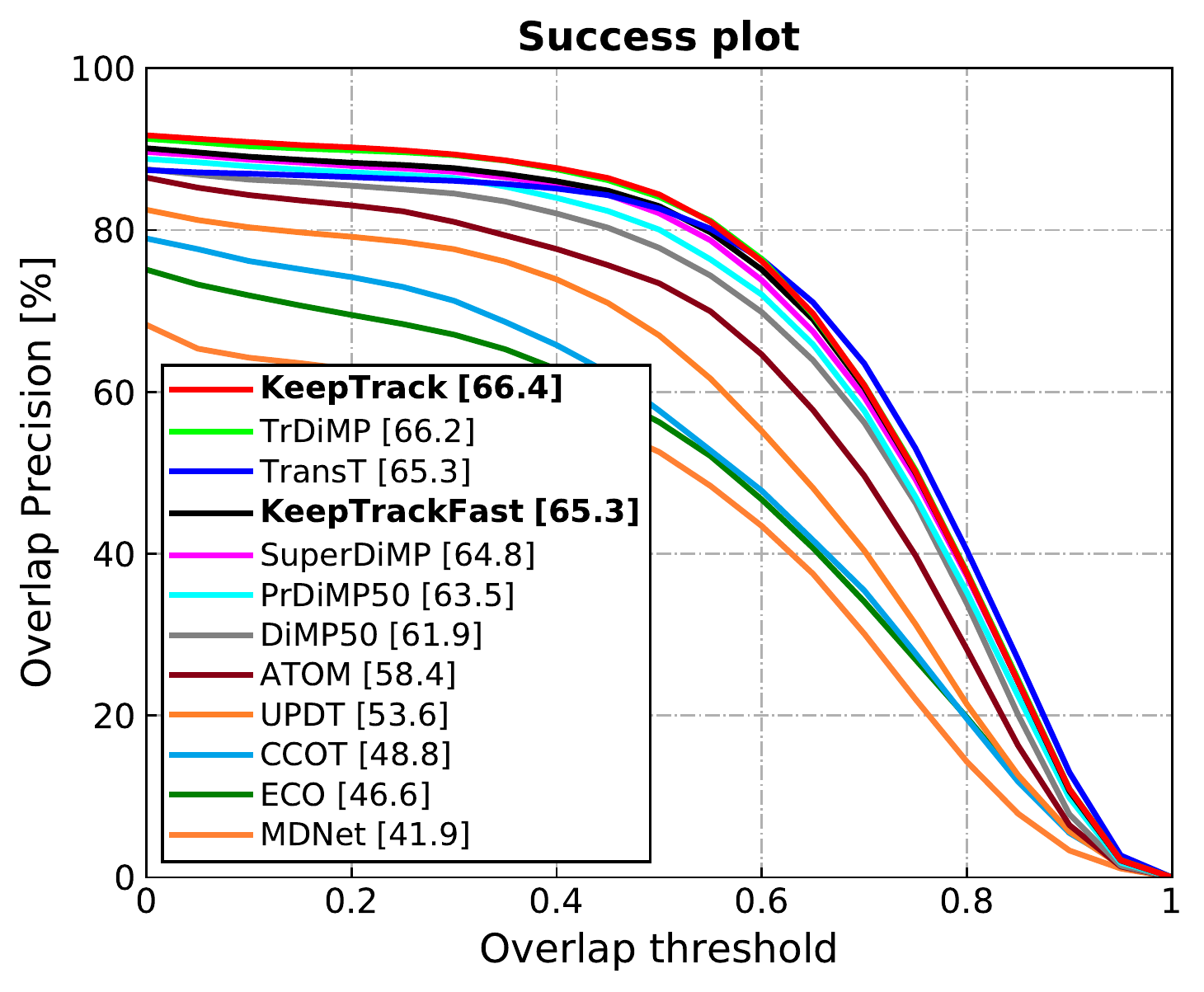}}%
	\caption{Success plots on the UAV123~\cite{Mueller_2016_ECCV_UAV123}, OTB-100~\cite{WU_2015_TPAMI_OTB} and NFS~\cite{Galoogahi_2017_ICCV_NFS} datasets in terms of overall AUC score, reported in the legend.}%
	\label{sup:fig:otb_ns_uav}%
\end{figure*}

%% file: supplementary/tables/otb_nfs_uav.tex
\begin{table*}[t]
	\centering
	\newcommand{\best}[1]{\textbf{\textcolor{red}{#1}}}
	\newcommand{\scnd}[1]{\textbf{\textcolor{blue}{#1}}}
	\newcommand{\dist}{\hspace{5pt}}%
	\resizebox{1.00\textwidth}{!}{%
        \begin{tabular}{l@{\dist}c@{\dist}c@{\dist}c@{\dist}c@{\dist}c@{\dist}c@{\dist}c@{\dist}c@{\dist}c@{\dist}c@{\dist}c@{\dist}c@{\dist}c@{\dist}c@{\dist}c@{\dist}c@{\dist}c@{\dist}c@{\dist}c@{\dist}c@{\dist}c@{\dist}c@{\dist}c@{\dist}c@{\dist}c@{\dist}c@{\dist}c@{\dist}c@{\dist}c@{\dist}c@{\dist}c@{\dist}c@{\dist}c@{\dist}}
        	\toprule
        	        & \textbf{Keep}  & \textbf{Keep}  & Tr   &        & Super       & Pr         & Siam  & STM   &      &      & Siam  &      &      & Retina      & FCOS &       &       \\
        	        & \textbf{Track} & \textbf{Track} & DiMP & TransT & DiMP        & DiMP       & R-CNN & Track & DiMP & KYS  & RPN++ & ATOM & UPDT & MAML        & MAML & Ocean & STN   \\
        	        & & \textbf{Fast} &\cite{Wang_2021_CVPR_TrDiMP} & \cite{Chen_2021_CVPR_TransT} &\cite{Danelljan_2019_github_pytracking} & \cite{Danelljan_2020_CVPR_PRDIMP} & \cite{Voigtlaender_2020_CVPR_SiamRCNN} & \cite{Fu_2021_CVPR_STMTrack} & \cite{Bhat_2019_ICCV_DIMP} & \cite{Bhat_2020_ECCV_KYS} & \cite{Li_2019_CVPR_SiamRPN++} & \cite{Danelljan_2019_CVPR_ATOM} & \cite{Bhat_2018_ECCV_UPDT} & \cite{Wang_2020_CVPR_MAML} & \cite{Wang_2020_CVPR_MAML} & \cite{Zhang_2020_ECCV_Ocean} & \cite{Liu_2020_ECCV_STN} \\
        	\midrule
        	UAV123 & \best{69.7} & \scnd{69.5} & 67.5        & 69.1 & 67.7 & 68.0 & 64.9 & 64.7        & 65.3 & --   & 61.3 & 64.2  & 54.5 & --   & --   & --   & 64.9  \\
        	OTB-100& 70.9        & 71.2        & 71.1        & 69.4 & 70.1 & 69.6 & 70.1 & \scnd{71.9} & 68.4 & 69.5 & 69.6 & 66.9  & 70.2 & 71.2 & 70.4 & 68.4 & 69.3  \\
        	NFS    & \best{66.4} & 65.3        & \scnd{66.2} & 65.7 & 64.8 & 63.5 & 63.9 & --          & 62.0 & 63.5 & --   & 58.4  & 53.7 & --   & --   & --   & --    \\\bottomrule
        	\toprule
        	& Auto  & Siam & Siam &      &       &        &             &     & Siam  &       &      & Siam & Siam &       &      & DaSiam &       \\
        	& Track & BAN  & CAR  & ECO  & DCFST & PG-NET & CRACT       & GCT & GAT   & CLNet & TLPG & AttN & FC++ & MDNet & CCTO & RPN    & ECOhc \\
        	& \cite{Li_2020_CVPR_AutoTrack} & \cite{Chen_2020_CVPR_SiamBAN} & \cite{Guo_2020_CVPR_SiamCAR} & \cite{Danelljan_2017_CVPR_ECO} & \cite{Zheng_2020_ECCV_DCFST} & \cite{Bingyan_2020_ECCV_PGNet} & \cite{Fan_2020_arxiv_CRACT} & \cite{Gao_2019_CVPR_GCT} & \cite{Guo_2020_arxiv_SiamGAT} & \cite{Dong_2020_ECCV_CLNet} & \cite{Li_2020_IJCAI_TLPG} & \cite{Yu_2020_CVPR_SiamAttN} & \cite{Xu_2020_AAAI_SiamFCpp} & \cite{Nam_2016_CVPR_MDNet} & \cite{Danelljan_2016_ECCV_CCTO} & \cite{Zhu_2018_ECCV_DaSiamRPN} & \cite{Danelljan_2017_CVPR_ECO}\\
        	\midrule
        	UAV123 & 67.1 & 63.1 & 61.4 & 53.2 & --   & --   & 66.4        & 50.8 & 64.6 & 63.3 & --   & 65.0 & --   & --   & 51.3 & 57.7 & 50.6 \\
        	OTB-100& --   & 69.6 & --   & 69.1 & 70.9 & 69.1 & \best{72.6} & 64.8 & 71.0 & --   & 69.8 & 71.2 & 68.3 & 67.8 & 68.2 & 65.8 & 64.3 \\
        	NFS    & --   & 59.4 & --   & 46.6 & 64.1 & --   & 62.5        & --   & --   & 54.3 & --   & --   & --   & 41.9 & 48.8 & --   & --   \\
        	\bottomrule
        \end{tabular}
	}
	\caption{Comparison with state-of-the-art on the OTB-100~\cite{WU_2015_TPAMI_OTB}, NFS~\cite{Galoogahi_2017_ICCV_NFS} and UAV123~\cite{Mueller_2016_ECCV_UAV123} datasets in terms of overall AUC score. The average value over 5 runs is reported for our approach.}
	\label{sup:tab:otb_nfs_uav}%
\end{table*}

%% file: supplementary/tables/vot2018lt.tex
\begin{table}[!b]
	\centering
	\newcommand{\best}[1]{\textbf{\textcolor{red}{#1}}}
	\newcommand{\scnd}[1]{\textbf{\textcolor{blue}{#1}}}
	\newcommand{\dist}{\hspace{10pt}}%
	\resizebox{1.00\columnwidth}{!}{%
        \begin{tabular}{l@{\dist}c@{\dist}c@{\dist}c@{\dist}c@{\dist}c@{\dist}c@{\dist}c@{\dist}c@{\dist}c@{\dist}c@{\dist}c@{\dist}}
        	\toprule
        	          & \textbf{Keep}  & \textbf{Keep}  &      & Siam  &       & Siam  & Super &      &      &          \\
        	          & \textbf{Track} & \textbf{Track} & LTMU & R-CNN & PGNet & RPN++ & DiMP  & SPLT & MBMD & DaSiamLT \\
        	          &                & \textbf{Fast}  & \cite{Dai_2020_CVPR_LTMU} & \cite{Voigtlaender_2020_CVPR_SiamRCNN} & \cite{Bingyan_2020_ECCV_PGNet} & \cite{Li_2019_CVPR_SiamRPN++} & \cite{Danelljan_2019_github_pytracking} & \cite{Yan_2019_ICCV_SPLT} & \cite{Zhang_2018_IJCV_MBMD} & \cite{Zhu_2018_ECCV_DaSiamRPN,Matej_2018_ECCVW_VOT2018}\\
        	\midrule
        	Precision          & \best{73.8} & 70.1 & \scnd{71.0} & --    & 67.9 & 64.9  & 64.3  & 63.3 & 63.4 & 62.7 \\
        	Recall             & \best{70.4} & 67.6 & \scnd{67.2} & --    & 61.0 & 60.9  & 61.0  & 60.0 & 58.8 & 58.8 \\
        	\textbf{F-Score}   & \best{72.0} & 68.8 & \scnd{69.0} & 66.8  & 64.2 & 62.9  & 62.2  & 61.6 & 61.0 & 60.7 \\\bottomrule
        \end{tabular}
	}
	\caption{Results on the VOT2018LT dataset~\cite{Matej_2018_ECCVW_VOT2018} in terms of F-Score, Precision and Recall.}
	\label{sup:tab:vot2018lt}%
\end{table}

%% file: supplementary/tables/uav_attributes.tex
\begin{table*}[t]
	\centering
	\newcommand{\best}[1]{\textbf{\textcolor{red}{#1}}}
	\newcommand{\scnd}[1]{\textbf{\textcolor{blue}{#1}}}
	\newcommand{\dist}{\hspace{5pt}}%
	\resizebox{1.00\textwidth}{!}{%
        \begin{tabular}{l@{\dist}c@{\dist}c@{\dist}c@{\dist}c@{\dist}c@{\dist}c@{\dist}c@{\dist}c@{\dist}c@{\dist}c@{\dist}c@{\dist}c@{\dist}|c@{\dist}}
        	\toprule
        	                        & Scale        & Aspect          & Low          & Fast      & Full      & Partial      &               & Background  & Illumination & Viewpoint  & Camera      & Similar     &        \\
        	                        & Variation    & Ratio Change    & Resolution   & Motion    & Occlusion & Occlusion    & Out-of-View   & Clutter     & Variation    & Change     & Motion      & Object      & Total  \\

        	\midrule
        	ATOM                    & 63.0        & 61.9           & 49.5        & 62.7        & 46.2        & 58.1        & 61.4          & 46.2        & 63.1        & 65.1        & 66.4        & 63.1        & 64.2 \\
            DiMP50                  & 63.8        & 62.8           & 50.9        & 62.7        & 47.5        & 59.7        & 61.8          & 48.9        & 63.9        & 65.2        & 66.9        & 62.9        & 65.3 \\
            STMTrack                & 63.9        & 64.2           & 46.4        & 62.2        & 48.9        & 58.0        & \scnd{68.2}   & 46.2        & 61.9        & 70.2        & 67.5        & 58.0        & 65.7 \\
            TrDiMP                  & 66.4        & 66.1           & 54.3        & 66.3        & 48.6        & 62.1        & 66.3          & 45.1        & 61.5        & 70.0        & 68.3        & 64.9        & 67.5 \\
            SuperDiMP               & 66.6        & 66.4           & 54.9        & 65.1        & 52.0        & 63.5        & 63.7          & 51.4        & 63.2        & 67.8        & 69.8        & 65.5        & 67.7 \\
            PrDiMP50                & 66.8        & 66.3           & 55.2        & 65.3        & 53.6        & 63.5        & 63.9          & 53.9        & 62.4        & 69.4        & 70.4        & 66.1        & 68.0 \\
            TransT                  & 68.0        & 66.3           & 55.6        & \scnd{67.4} & 48.4        & 63.2        & \best{69.1}   & 44.1        & 62.6        & \best{71.8} & 70.5        & 65.3        & 69.1 \\
            \textbf{KeepTrackFast}  & \scnd{68.4} & \scnd{68.8}    & \best{57.3} & 67.2        & \scnd{55.4} & \best{66.0} & 65.9          & \best{55.4} & \best{65.6} & 70.3        & \scnd{71.2} & \best{67.9} & \scnd{69.5} \\
            \textbf{KeepTrack}      & \best{68.7} & \best{68.9}    & \scnd{57.0} & \best{68.0} & \best{55.5} & \scnd{65.8} & 66.8          & \scnd{55.2} & \scnd{65.4} & \scnd{70.4} & \best{71.8} & \scnd{67.2} & \best{69.7} \\
            \bottomrule
        \end{tabular}
	}
	\caption{UAV123 attribute-based analysis in terms of AUC score. Each column corresponds to the results computed on all sequences in the dataset with the corresponding attribute.}
	\label{sup:tab:uav_attributes}%
\end{table*}

%% file: supplementary/tables/lasot_attributes.tex
\begin{table*}[t]
	\centering
	\newcommand{\best}[1]{\textbf{\textcolor{red}{#1}}}
	\newcommand{\scnd}[1]{\textbf{\textcolor{blue}{#1}}}
	\newcommand{\dist}{\hspace{5pt}}%
	\resizebox{1.00\textwidth}{!}{%
        \begin{tabular}{l@{\dist}c@{\dist}c@{\dist}c@{\dist}c@{\dist}c@{\dist}c@{\dist}c@{\dist}c@{\dist}c@{\dist}c@{\dist}c@{\dist}c@{\dist}c@{\dist}c@{\dist}|c@{\dist}}
        	\toprule
        	                       & Illumination & Partial     &                & Motion         & Camera      &             & Background & Viewpoint   & Scale        & Full        & Fast        &             & Low         & Aspect        &       \\
        	                       & Variation    & Occlusion   & Deformation    & Blur           & Motion      & Rotation    & Clutter    & Change      & Variation    & Occlusion   & Motion      & Out-of-View & Resolution  & Ration Change & Total \\
 
        	\midrule
            LTMU                   & 56.5         & 54.0        & 57.2           & 55.8           & 61.6        & 55.1        & 49.9        & 56.7        & 57.1        & 49.9        & 44.0        & 52.7        & 51.4        & 55.1          & 57.2 \\
            PrDiMP50               & 63.7         & 56.9        & 60.8           & 57.9           & 64.2        & 58.1        & 54.3        & 59.2        & 59.4        & 51.3        & 48.4        & 55.3        & 53.5        & 58.6          & 59.8 \\
            STMTrack               & 65.2         & 57.1        & 64.0           & 55.3           & 63.3        & 60.1        & 54.1        & 58.2        & 60.6        & 47.8        & 42.4        & 51.9        & 50.3        & 58.8          & 60.6 \\
            SuperDiMP              & 67.8         & 59.7        & 63.4           & 62.0           & 68.0        & 61.4        & 57.3        & 63.4        & 62.9        & 54.1        & 50.7        & 59.0        & 56.4        & 61.6          & 63.1 \\
            TrDiMP                 & 67.5         & 61.1        & 64.4           & 62.4           & 68.1        & 62.4        & 58.9        & 62.8        & 63.4        & 56.4        & 53.0        & 60.7        & 58.1        & 62.3          & 63.9 \\
            Siam R-CNN             & 64.6         & 62.2        & 65.2           & 63.1           & 68.2        & 64.1        & 54.2        & 65.3        & 64.5        & 55.3        & 51.5        & 62.2        & 57.1        & 63.4          & 64.8 \\
            TransT                 & 65.2         & 62.0        & \best{67.0}    & 63.0           & 67.2        & 64.3        & 57.9        & 61.7        & 64.6        & 55.3        & 51.0        & 58.2        & 56.4        & 63.2          & 64.9 \\
            AlphaRefine            & 69.4         & 62.3        & 66.3           & \scnd{65.2}    & 70.0        & 63.9        & 58.8        & 63.1        & 65.4        & 57.4        & 53.6        & 61.1        & 58.6        & 64.1          & 65.3 \\
            \textbf{KeepTrackFast} & \best{70.1}  & \scnd{63.8} & 66.2           & 65.0           & \scnd{70.7} & \scnd{65.1} & \scnd{60.1} & \best{67.6} & \scnd{66.6} & \scnd{59.2} & \scnd{57.1} & \scnd{63.4} & \best{62.0} & \scnd{65.6}   & \scnd{66.8} \\
            \textbf{KeepTrack}     & \scnd{69.7}  & \best{64.1} & \best{67.0}    & \best{66.7}    & \best{71.0} & \best{65.3} & \best{61.2} & \scnd{66.9} & \best{66.8} & \best{60.1} & \best{57.7} & \best{64.1} & \best{62.0} & \best{65.9}   & \best{67.1} \\
            \bottomrule
        \end{tabular}
	}
	\caption{LaSOT attribute-based analysis. Each column corresponds to the results computed on all sequences in the dataset with the corresponding attribute.}
	\label{sup:tab:lasot_attributes}%
\end{table*}